\documentclass[10pt,twocolumn,letterpaper]{article}

\usepackage[pagenumbers]{cvpr} 

\usepackage[T1]{fontenc}

\usepackage{amsthm}
\theoremstyle{plain}

\theoremstyle{definition}

\usepackage{algpseudocode}

\usepackage[skins,breakable]{tcolorbox}
\newtcolorbox{promptbox}[1]{breakable, enhanced, colback=black!3, colframe=black!50,
  boxrule=0.4pt, arc=2pt, left=4pt, right=4pt, top=3pt, bottom=3pt, boxsep=1pt,
  fonttitle=\bfseries\scriptsize, coltitle=black, colbacktitle=black!12, title={#1},
  fontupper=\scriptsize\ttfamily, before upper=\raggedright}

\usepackage{multirow}
\usepackage{colortbl}

\definecolor{dgreen}{rgb}{0.0,0.5,0.0}
\definecolor{dred}{rgb}{0.75,0.0,0.0}
\newcommand{\gain}[1]{{\tiny\textcolor{dgreen}{#1}}}
\newcommand{\loss}[1]{{\tiny\textcolor{dred}{#1}}}

\newcommand{\trainsize}{1{,}000}
\definecolor{methodcolor}{HTML}{50164A}
\colorlet{oursbg}{methodcolor!8}
\newcommand{\method}{\textcolor{methodcolor}{\textup{\textsc{OracleZoom}}}}
\newcommand{\stepnum}[1]{%
\smash{%
\tikz[baseline={([yshift=-0.3em]char.center)}]{
\node[
    circle,
    draw=stepdeep#1,
    fill=steplight#1,
    text=stepdeep#1,
    inner sep=0pt,
    minimum size=0.95em,
    line width=0.7pt,
    font=\fontsize{6}{6}\selectfont\bfseries
] (char) {#1};}%
}}

\usepackage{xcolor}
\definecolor{algocol}{rgb}{0.74,0.18,0.05}
\definecolor{algcomment}{HTML}{377E22}
\definecolor{stepdeep1}{HTML}{2E7D4F}\definecolor{steplight1}{HTML}{EEF7F1}
\definecolor{stepdeep2}{HTML}{B71C1C}\definecolor{steplight2}{HTML}{FDF0F0}
\definecolor{stepdeep3}{HTML}{1F5FA8}\definecolor{steplight3}{HTML}{EEF3FB}
\definecolor{stepdeep4}{HTML}{B07800}\definecolor{steplight4}{HTML}{FFF7E6}
\definecolor{stepdeep5}{HTML}{6A3D9A}\definecolor{steplight5}{HTML}{F5EEFA}
\usepackage[linesnumbered,vlined,ruled,fillcomment]{algorithm2e}
\SetAlFnt{\footnotesize}
\SetAlgoNlRelativeSize{-2}
\SetAlCapFnt{\footnotesize}

\SetCommentSty{mycommfont}
\SetKwComment{Comment}{$\triangleright$\ }{}

\definecolor{phasebg}{HTML}{E8EEF7}

\usepackage{placeins}   

\makeatletter
\patchcmd{\@algocf@start}{-1.5em}{0pt}{}{}
\makeatother

\DontPrintSemicolon

\definecolor{cvprblue}{rgb}{0.21,0.49,0.74}
\usepackage[pagebackref,breaklinks,colorlinks,allcolors=cvprblue]{hyperref}

\usepackage{makecell}  
\usepackage[table]{xcolor}
\PassOptionsToPackage{most}{tcolorbox}
\definecolor{tzBlueHeader2}{RGB}{105,185,225}
\definecolor{tzBlueBorder}{RGB}{115,190,225}
\definecolor{tzBlueFill}{RGB}{232,246,252}

\usepackage{pifont}              

\newcommand{\panelgraphic}[3][3.0cm]{%
  \IfFileExists{#2}
    {\includegraphics[width=\linewidth]{#2}}
    {\fbox{%
      \parbox[c][#1][c]{0.94\linewidth}{%
        \centering\scriptsize #3%
      }%
    }}%
}

\newtcolorbox{jsonbox}{
  enhanced,
  breakable,
  colback=white,
  colframe=tzBlueBorder!60!white,
  boxrule=0.6pt,
  arc=3pt,
  left=6pt, right=6pt, top=4pt, bottom=4pt,
}

\def\confName{CVPR}
\def\confYear{2027}

\title{\method{}: Reference-Constrained Recursive Image Super-Resolution Beyond Ground Truth}

\author{Shubhashis Roy Dipta$^{1}$ \qquad Nobin Sarwar$^{1}$ \qquad Md Rizwan Parvez$^{2}$\\
$^{1}$University of Maryland, Baltimore County\\
$^{2}$Qatar Computing Research Institute (QCRI)\\
{\tt\small \{sroydip1, sms2\}@umbc.edu \quad mparvez@hbku.edu.qa}\\
{\small Project page: \url{https://dipta007.github.io/OracleZoom/}}
}

\begin{document}
\twocolumn[{%
\renewcommand\twocolumn[1][]{#1}%
    \maketitle
    \centering
    \includegraphics[width=0.97\linewidth]{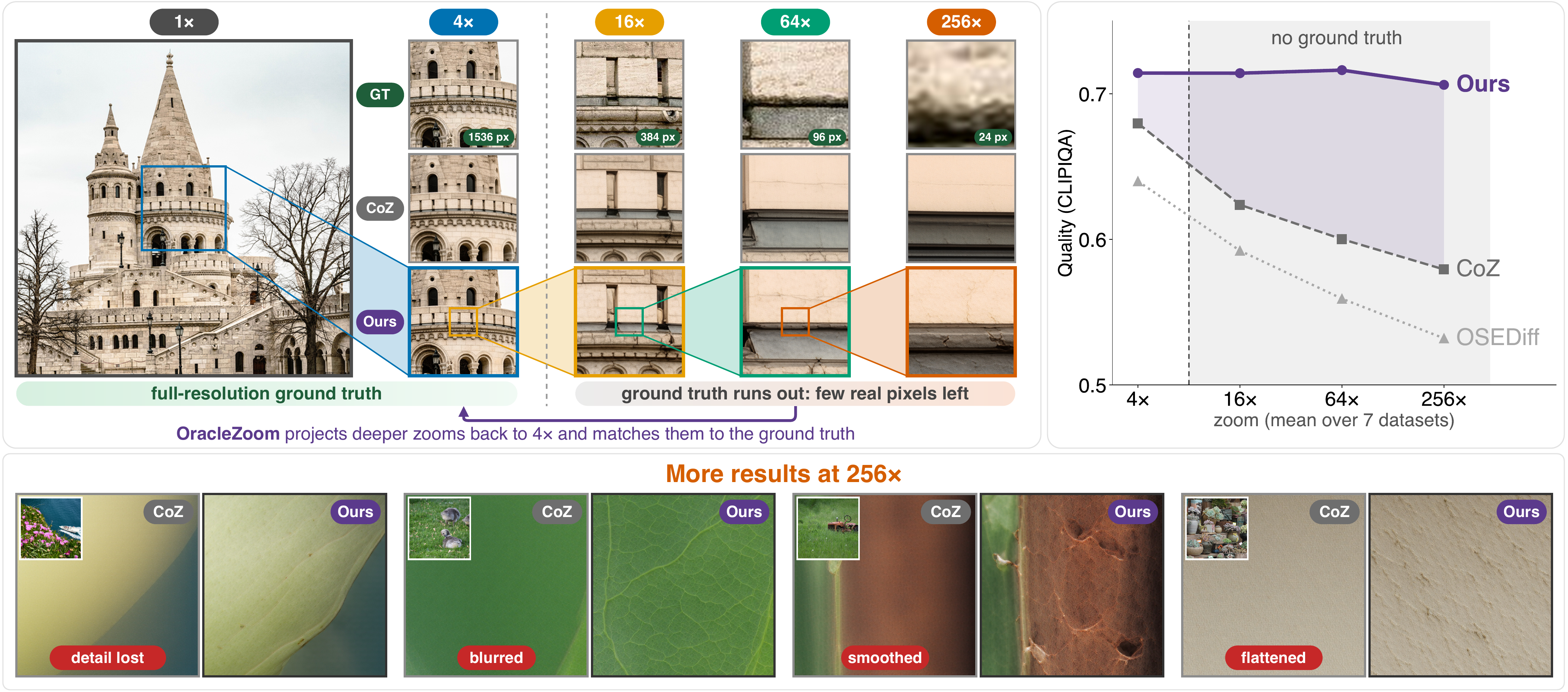}
    \vspace{-6pt}
    \captionof{figure}
    {
    \textbf{\method{} keeps extreme zooms sharp after ground truth runs out.}
    Top left: full-resolution ground truth stops at $4{\times}$. At $64{\times}$ and $256{\times}$ there is no ground truth: the GT row only enlarges $96$ and $24$ real pixels.
    During training, \method{} projects the $16{\times}$ zoom back to $4{\times}$ and matches it to this ground truth.
    At $16{\times}$, the GT still shows the stone joints.
    \method{} keeps sharp edges at deeper zooms, while CoZ~\cite{kim2025chain} smooths the wall flat.
    Top right: mean CLIPIQA over seven test sets.
    Bottom: more $256{\times}$ outputs.
}
    \label{fig:teaser}
    \vspace{8pt}
}
]

\begin{abstract}
Recursive image Super-Resolution (SR) reaches extreme magnification by feeding each prediction back into the same model, like zooming into an image again and again.
Real ground truth for deep zooms is impractical to capture (a full $256\times$ target would be a $52$\,GB image), so predictions at deeper zooms have no supervision.
Yet the last ground-truth image still contains the region that every deeper zoom magnifies.
\method{} turns this region into a training signal.
Inspired by on-policy self-distillation, the model trains on its own prediction for one recursive step ($4\times\rightarrow16\times$), and a cross-scale loss projects each $16\times$ prediction back to the last ground truth at $4\times$.
This projection can verify only coarse content, so a no-reference quality objective guides the remaining fine detail, and a latent prior stops this objective from adding hallucinated artifacts.
On seven datasets, \method{} achieves the highest CLIPIQA at every zoom depth (mean $0.713$ vs.\ $0.621$ for Chain-of-Zoom), with the largest gains at the deepest zooms, even though training stops at $16\times$ and optimizes a different quality metric.
Pooled over two ground-truth-anchored judges, this added detail comes at the same hallucination rate as Chain-of-Zoom at $16\times$ and a lower rate at $4\times$, and removing the cross-scale loss raises hallucination at $16\times$.
\vspace{-16pt}
\end{abstract}

\section{Introduction}
\label{sec:intro}
Single-image Super-Resolution (SR) reconstructs a high-resolution image from a low-resolution observation.
Recent generative SR methods~\cite{wang2024stablesr, gao2023implicit, wu2024seesr, yu2024supir, wang2024sinsr, wu2024one} use strong diffusion models~\cite{dhariwal2021diffusion} to recover more realistic high-frequency detail than regression-based reconstruction~\cite{lim2017edsr, liang2021swinir, zhang2024hierarchical}.
Recursive SR goes further. It applies an SR model again and again, and each prediction becomes the input to the next zoom.
A viewer can then keep zooming into a photograph far beyond the resolution of the camera, for example to inspect a texture or a distant object.
Chain-of-Zoom (CoZ)~\cite{kim2025chain} works in this way and reaches $256{\times}$ zoom with guidance from a Vision-Language Model (VLM).

The lack of ground truth is the main obstacle to training a recursive SR model.
For a $512{\times}512$ input, a full $256{\times}$ target would be a $131072{\times}131072$ image of about $52$ GB.
Capturing pixel-aligned photographs at this density for a whole training set is impractical.
Ground truth is therefore available only for the earliest zooms.
In \cref{fig:teaser} (top left), the $4{\times}$ view still has full-resolution ground truth. At $64{\times}$ and $256{\times}$ there is no ground truth: the GT row only enlarges the $96$ and $24$ real pixels left.
Later zooms must create finer detail without a visual target, and quality drops with depth.
The mean CLIPIQA of CoZ falls from $0.680$ at $4{\times}$ to $0.579$ at $256{\times}$ (\cref{fig:teaser}, top right). Its deepest zooms lose, blur, or flatten detail (\cref{fig:teaser}, bottom).
We call this lack of ground truth the \emph{supervision gap}: the model relies more on its own predictions where no ground truth can check them.

Progressive and arbitrary-scale SR methods~\cite{lai2017lapsrn, chen2021liif} leave this gap open: they supervise every output scale with ground truth, so they get no training signal beyond the last ground-truth scale.
The VLM prompts of CoZ give useful semantic guidance~\cite{kim2025chain}, but text cannot check whether the generated textures and structures match the observed image.
Other methods project a prediction to a coarser scale and compare it with an observed image~\cite{irani1991improving, guo2020closed, zhou2022aris}.
They compare within a single SR step, either with the input of that step or with the available ground truth.
In a recursive chain, however, the input of every zoom after the first is itself a prediction.

We call the last scale with ground truth the \emph{supervision boundary}.
To close the supervision gap, we introduce \method{}, a training method that brings ground truth into the zooms beyond this boundary.
Its key observation is that the last ground-truth image still contains the region that every deeper zoom magnifies.
\method{} trains the SR model on its own recursive predictions: each predicted image becomes the input to the next zoom, as at inference.
This training is inspired by On-Policy Self-Distillation (OPSD)~\cite{zhao2026selfdistilled}, its recent visual and generative extensions~\cite{yuan2026vision, li2026visual, liu2026opsdv, zhou2026diffusionopsd, zhao2026mtopsd}, and on-policy distillation for VLMs~\cite{bousselham2026vold}.
A \emph{cross-scale consistency} loss projects each deeper prediction back to the resolution of the boundary (purple arrow in \cref{fig:teaser}).
It then compares the projected prediction with the region of the last ground truth that the prediction magnifies.
This reference is real ground truth, not the input of the zoom.
The loss therefore ties deeper zooms to the real scene without any deeper target.

Projection can verify only coarse content, because many high-resolution predictions project to the same lower-resolution image~\cite{wang2023ddnm}.
To guide the finer detail, we use a frozen no-reference quality model, TOPIQ-NR~\cite{chen2024topiq}, and we evaluate with four other metrics (\cref{tab:main}).
Optimizing quality alone favors sharp patterns that the image does not support~\cite{blau2018perception}.
A latent prior therefore keeps the latents of the fine-tuned model close to those of the pretrained SR model.
Without this prior, nearly every $16{\times}$ zoom is flagged as hallucinated (\cref{sec:ablations}).
Consistency with an exponential moving average (EMA) copy of the model slightly improves fidelity at the supervision boundary.
We fine-tune only a $7.1$M-parameter LoRA adapter~\cite{hu2022lora} on the SR model of CoZ, using $1{,}000$ images and a single recursive step ($4{\times}{\rightarrow}16{\times}$). Only inference reaches $64{\times}$ and $256{\times}$.

On seven datasets, \method{} achieves the highest CLIPIQA~\cite{wang2023clipiqa} at every zoom depth and stays above $0.70$ even at $256{\times}$ (\cref{fig:teaser}, top right).
In the castle example of \cref{fig:teaser}, the ground truth at $16{\times}$ still shows the joints between the stones. \method{} keeps sharp edges at every deeper zoom, while the zooms of CoZ smooth the wall flat.
Two VLM judges from different vendors compare each zoom with the native ground-truth region that it magnifies.
Up to the $16{\times}$ zoom level, \method{} hallucinates no more often than CoZ.
In a blind study, two human raters also preferred \method{} in $80\%$ of the non-tie votes (supplementary material).
Without the cross-scale loss, the hallucination rate at $16{\times}$ rises from $0.45$ to $0.59$.
Beyond $16{\times}$, this check becomes unreliable for every method.
On the high-resolution test sets, the ground-truth region holds about $256$ native pixels at $16{\times}$ but only about $64$ at $64{\times}$.
That is too few for the judges to tell real photos from model outputs.

\vspace{3pt}

\noindent \textbf{Our contributions are:}
\begin{itemize}[nosep]
    \item \textbf{Reference-Constrained Recursion Beyond Ground Truth.}
    To our knowledge, \method{} is the first method that fine-tunes a recursive SR model on its own predictions beyond the last ground-truth scale.
    Cross-scale consistency ties these predictions back to that ground truth.
    \item \textbf{Supervision for Verifiable and Unverifiable Detail.}
    Cross-scale consistency constrains the content that the last ground truth can still verify.
    Quality guidance supplies the finer detail, and a latent prior keeps this detail from turning into artifacts.
    \item \textbf{More Detail at the Same Hallucination Rate.}
    On seven datasets, \method{} achieves the highest CLIPIQA at every zoom depth.
    Up to $16{\times}$, it hallucinates no more often than CoZ and matches the last ground truth most closely among the compared methods.
\end{itemize}

\begin{figure*}[t]
    \centering
    \includegraphics[width=\linewidth,height=0.87\textheight,keepaspectratio]{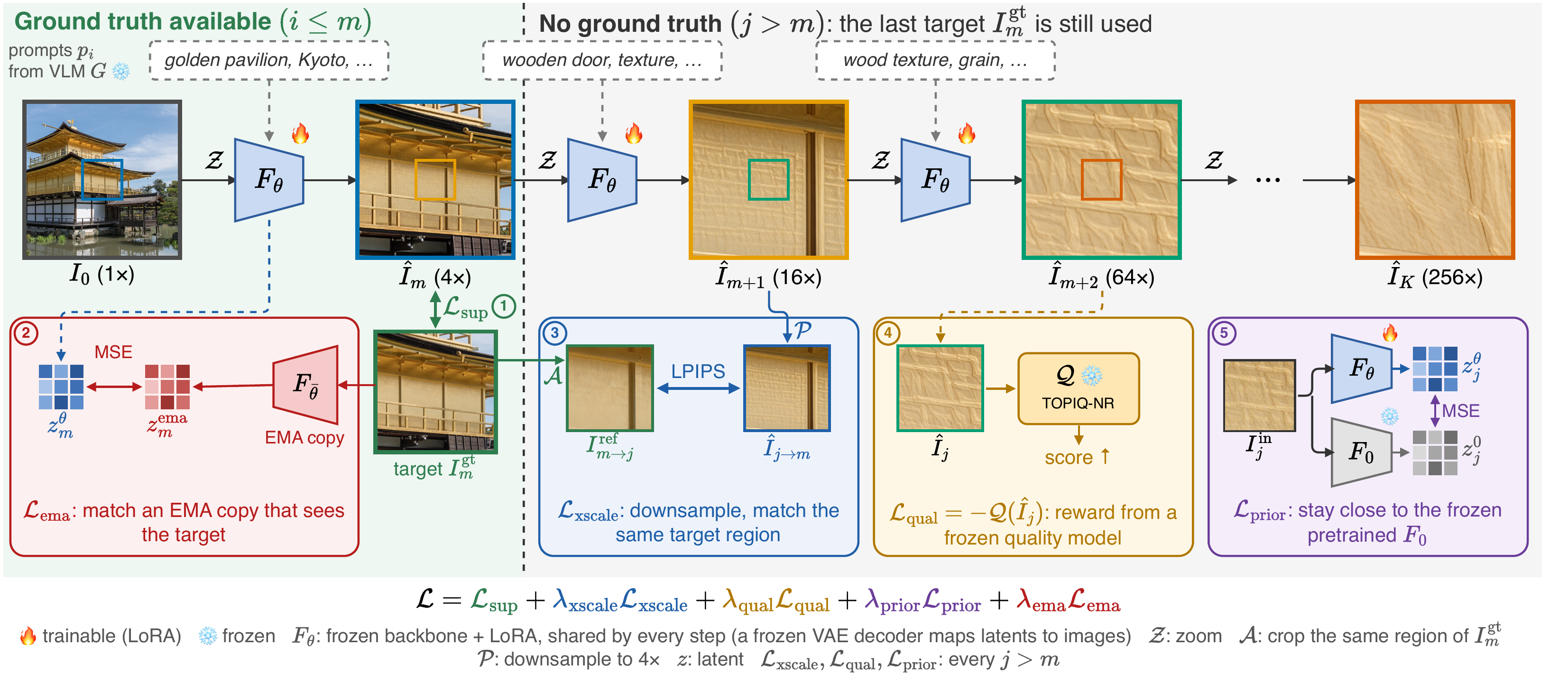}
    \caption{
    \textbf{Overview of \method{}.}
    One SR model $F_\theta$ zooms the image recursively, with text prompts from a frozen VLM $G$.
    Up to step $m$, ground truth supervises the predictions ($\mathcal{L}_{\mathrm{sup}}$, $\mathcal{L}_{\mathrm{ema}}$).
    Beyond step $m$, each prediction is matched to the same region of $I_m^{\mathrm{gt}}$ ($\mathcal{L}_{\mathrm{xscale}}$), a frozen quality model guides the remaining detail ($\mathcal{L}_{\mathrm{qual}}$), and the frozen base model $F_0$ keeps this detail close to its output ($\mathcal{L}_{\mathrm{prior}}$).
    Our experiments train with $m{=}1$ and $K{=}2$ ($16\times$) and run inference to $256\times$.
}
\label{fig:method}
\vspace{-10pt}
\end{figure*}

\section{Related Work}
\label{sec:related_work}

\noindent\textbf{Super-Resolution and Extreme Magnification.} 
Single-image SR has progressed from regression and adversarial reconstruction~\cite{lim2017edsr, wang2018esrgan, liang2021swinir} to blind and real-world restoration that handles unknown degradations~\cite{zhang2021bsrgan, wang2021realesrgan, wang2021dasr}. 
More recently, diffusion and large generative priors let SR methods synthesize richer perceptual detail~\cite{fei2023gdp, wang2024stablesr, yue2023resshift, lin2024diffbir, wu2024seesr, yu2024supir, wang2024sinsr, wu2024one, yang2024pasd, sun2024coser}. 
Parallel work supports progressive or arbitrary-scale SR through pyramidal reconstruction and continuous image representations~\cite{lai2017lapsrn, hu2019metasr, chen2021liif, lee2022lte, cao2023ciaosr, chen2023clit, wang2023scalesr}. 
These methods supervise every output scale with ground truth. Our setting instead reuses predictions recursively after ground-truth supervision ends.
The recent Chain-of-Zoom (CoZ)~\cite{kim2025chain} addresses this extreme-magnification setting. It applies a fixed-scale SR model recursively and uses multi-scale VLM guidance to steer each zoom step, so deep zooms receive text guidance but no visual supervision.
Other recent methods reach large scales by applying one SR model cyclically or in a cascade~\cite{bang2025casarbi, guo2025sbca, guo2026casr}, or by fine-tuning on real close-up photographs~\cite{ma2025ultrazoom, huynh2026microzoom}, but none of them trains on its own predictions from earlier zoom steps.
Methods outside recursive SR also use training signals beyond the supervision boundary. ARIS~\cite{zhou2022aris} downsamples outputs at unseen scales to the available ground truth, and ZSSR~\cite{shocher2018zssr} adds its own intermediate outputs to its training set.
Iterative back-projection~\cite{irani1991improving} and dual regression~\cite{guo2020closed} also project a prediction to a coarser scale and compare it with an observed image.
Zoom stacks~\cite{wang2024powers} and null-space restoration~\cite{wang2023ddnm} apply this comparison at inference with a frozen model.
We use this comparison as a training loss along a recursive chain: each deeper prediction comes from the model's own previous zoom, and the last ground truth still guides it.

\noindent\textbf{Learning Beyond Direct Supervision.} 
When targets are unavailable, indirect signals include no-reference quality estimators~\cite{ke2021musiq, yang2022maniqa, wang2023clipiqa, chen2024topiq} and consistency with slowly updated EMA teachers~\cite{tarvainen2017meanteacher, caron2021dino}.
OPSD~\cite{zhao2026selfdistilled} trains a student on its own trajectories, with a teacher that sees privileged information, which reduces mismatch between inputs at training and at inference.
We adopt this perspective for recursive SR. The SR model is trained on images produced by preceding zooms, and ground truth is used only during training.
Training through a model's own intermediate outputs also appears in progressive SR~\cite{lai2019mslapsrn}, where every level has ground truth, in scale-wise image generation~\cite{zhou2025sar}, in autoregressive video generation~\cite{huang2025selfforcing, cui2026selfforcingpp}, in iterative image-to-image refinement~\cite{hassani2026rrfc}, and in multi-turn image editing~\cite{zhao2026mtopsd, dorfman2026acedit}. These methods rely on targets or a teacher at the same resolution as the prediction. In contrast, we add a cross-scale reference: each deeper prediction is projected back to a real ground-truth image.
No-reference scores can rise without visual improvement~\cite{antsiferova2024comparing} and can reward plausible detail unsupported by the input~\cite{blau2018perception, ren2025hallucscore, wu2025dp2osr}, and VLM-based scores measure such hallucination~\cite{ryou2026vha}. Recent SR methods bound such rewards with ground truth or score outputs against the low-resolution input~\cite{fan2026birdsr, fei2026lucidnft}, and K2N~\cite{fang2026k2n} generates only the finer scales over coarse scales taken from the input. We combine quality guidance with a latent prior from the pretrained model, as ACPO~\cite{yang2026acpo} does for diffusion image generation, and apply this pairing along a recursive SR chain.

\vspace{-4mm}
\section{Method}
\label{sec:method}
\method{} trains one SR model on its own zoom chain and uses the last ground truth as a reference for every deeper zoom (\cref{fig:method}).

\subsection{Problem Setup}
\label{sec:problem_setup}

Given a low-resolution image $I_0$, we construct a recursive zoom sequence using cumulative magnification factors $s_0 ( {=} 1 ) {<} s_1 {<} \cdots {<} s_K$. 
The relative magnification (\emph{zoom}) at step $i$ is $\gamma_i {=} s_i/s_{i-1}$. Let $\mathcal{Z}_{\gamma_i}$ denote the zoom operator at step $i$. It takes the center crop of side $H/\gamma_i$ from an $H{\times}H$ image ($H{=}512$) and resizes it back to $H{\times}H$ with bicubic interpolation.
$\mathcal{Z}_{\gamma_i}$ only selects and resamples the region of interest and adds no detail. The SR input is
\begin{equation}
    I_i^{\mathrm{in}} = \mathcal{Z}_{\gamma_i} \left( \hat I_{i-1} \right), \qquad \hat I_0 {=} I_0.
    \label{eq:zoom_input}
\end{equation}
At each step, a frozen VLM prompter $G$ receives the previous image and its zoomed crop. It returns a short text prompt for the crop (details in the supplementary material):
\begin{equation}
    p_i = G \left( \hat I_{i-1}, I_i^{\mathrm{in}} \right), \qquad i {\geq} 1.
    \label{eq:recursive_prompt}
\end{equation}
The prompt is text only, so no gradient flows through $G$.
Let $F_{\theta}$ denote the latent SR model. Its pretrained backbone remains frozen, and $\theta$ denotes the trainable LoRA adapter parameters~\cite{hu2022lora}.
Given $I_i^{\mathrm{in}}$ and $p_i$, the model produces the latent prediction $z_i^{\theta}$. 
A frozen VAE decoder $D$ then maps the latent $z_i^{\theta}$ to image space:
\begin{equation}
    z_i^{\theta} = F_{\theta} \left( I_i^{\mathrm{in}}, p_i \right), \qquad \hat I_i = D \left( z_i^{\theta} \right).
    \label{eq:sr_decode}
\end{equation}
Applying this recursively produces the SR sequence
\begin{equation}
    I_0 \rightarrow \hat I_1 \rightarrow \hat I_2 \rightarrow \cdots \rightarrow \hat I_K.
    \label{eq:recursive_sequence}
\end{equation}
Ground truth $I_i^{\mathrm{gt}}$ is available only for $i \leq m$, so $s_m$ is the last scale with ground truth, the supervision boundary.
Our goal is an SR model that stays faithful to the available targets for $i \leq m$ and to the verifiable content of $I_m^{\mathrm{gt}}$ at deeper scales $i > m$, while it adds realistic detail.
The supervision gap appears where recursion continues but ground truth no longer exists. We reuse the last target $I_m^{\mathrm{gt}}$ as a reference beyond $s_m$.

\subsection{Training Objectives}
\label{sec:oraclezoom_method}

We propose \method{} to address this gap with five complementary objectives:
\stepnum{1} direct supervision keeps predictions faithful at target-available scales,
\stepnum{2} EMA consistency adds a second target at the supervision boundary,
\stepnum{3} cross-scale consistency carries verifiable ground-truth information deeper into the recursion,
\stepnum{4} quality guidance encourages fine-scale detail that $I_m^{\mathrm{gt}}$ cannot resolve, and
\stepnum{5} a latent prior keeps this detail close to the pretrained SR model (\cref{fig:method}).
Beyond $s_m$, these objectives treat each prediction as two parts: content that $I_m^{\mathrm{gt}}$ can still verify, and finer detail that it cannot verify.

\begin{algorithm}[t]
    \newcommand{\algW}{\dimexpr\columnwidth-3.9pt\relax}
    \newcommand{\algL}[2]{{\color{stepdeep#1}\mathcal{L}_{\mathrm{#2}}}}
    \newcommand{\algN}[1]{\makebox[6pt][r]{\fontsize{5.5}{6}\selectfont\color{black!45}#1}}
    \newcommand{\algNote}[1]{{\scriptsize\color{black!58}#1}}
    \newcommand{\algEnd}[1]{\makebox[42pt][r]{\algNote{#1}}}
    \newcommand{\algSpine}[4]{\multirow{-#1}{*}[#3]{\hspace{#4}\rotatebox[origin=c]{90}{\fontsize{5.5}{6}\selectfont\bfseries #2}}}
    \newcommand{\algBar}[2]{\rowcolor{methodcolor!26}\multicolumn{4}{@{}l@{}}{\hspace{4pt}\makebox[\dimexpr\algW-8pt\relax][l]{\color{methodcolor}#1\hfill{\scriptsize #2}}\hspace{4pt}}}
    \newcommand{\algHead}[1]{\multicolumn{4}{@{}l@{}}{\hspace{4pt}#1}}
    \footnotesize
    \setlength{\tabcolsep}{2pt}\renewcommand{\arraystretch}{1.17}
    \hspace*{\dimexpr-\algomargin+1.95pt\relax}\begin{tabular}{@{}p{6pt}lp{\dimexpr\algW-66pt\relax}l@{}}
        \algHead{\textbf{Input:} image $I_0$, targets $\{I_i^{\mathrm{gt}}\}_{i=1}^{m}$, scales $\{s_i\}_{i=0}^{K}$} \\
        \algHead{\textbf{Output:} adapter $\theta$, recursive predictions $\{\hat I_i\}_{i=1}^{K}$} \\
        \algBar{\textbf{Training}\hspace{8pt}$\bar\theta\gets\theta$}{repeat until early stopping} \\
        & \algN{1} & Run the zoom chain: $\{\hat I_i,z_i^{\theta},p_i\}_{i=1}^{K}$ & \algEnd{Eqs.~\eqref{eq:zoom_input}--\eqref{eq:sr_decode}} \\
        \rowcolor{steplight1}\cellcolor{stepdeep1!30} & \algN{2} & \stepnum{1}\hspace{3pt}$\algL{1}{sup}\gets\operatorname{LPIPS}(\hat I_i,I_i^{\mathrm{gt}})$ & \algEnd{$i{\leq}m$, Eq.~\eqref{eq:supervised_loss}} \\
        \rowcolor{steplight2}\cellcolor{stepdeep1!30}\algSpine{2}{\color{stepdeep1!75!black}GT}{-0.06pt}{1.02pt} & \algN{3} & \stepnum{2}\hspace{3pt}$\algL{2}{ema}\gets\operatorname{MSE}(z_m^{\theta},\operatorname{sg}[F_{\bar\theta}(I_m^{\mathrm{gt}},p_m)])$ & \algEnd{$s_m$, Eq.~\eqref{eq:ema_loss}} \\
        \rowcolor{stepdeep3!16}\cellcolor{black!22} & \algN{4} & \stepnum{3}\hspace{3pt}$\algL{3}{xscale}\gets\operatorname{LPIPS}(\mathcal{P}_{j\rightarrow m}(\hat I_j),\mathcal{A}_{m\rightarrow j}(I_m^{\mathrm{gt}}))$ & \algEnd{\textbf{\color{stepdeep3}core}, Eq.~\eqref{eq:cross_scale_loss}} \\
        \rowcolor{steplight4}\cellcolor{black!22} & \algN{5} & \stepnum{4}\hspace{3pt}$\algL{4}{qual}\gets-\mathcal{Q}(\hat I_j)$ & \algEnd{$j{>}m$, Eq.~\eqref{eq:quality_loss}} \\
        \rowcolor{steplight5}\cellcolor{black!22}\algSpine{3}{\color{black!75}No GT}{-0.24pt}{1.02pt} & \algN{6} & \stepnum{5}\hspace{3pt}$\algL{5}{prior}\gets\operatorname{MSE}(z_j^{\theta},\operatorname{sg}[F_0(I_j^{\mathrm{in}},p_j)])$ & \algEnd{$j{>}m$, Eq.~\eqref{eq:prior_loss}} \\
        & \algN{7} & Update $\theta$ with $\nabla_{\theta}\mathcal{L}$; $\bar\theta\gets\mu\bar\theta+(1-\mu)\theta$ & \algEnd{Eq.~\eqref{eq:total_objective}} \\
        \algBar{\textbf{Inference}\hspace{8pt}$\hat I_0\gets I_0$}{no ground truth, no training-only branches} \\
        & \algN{8} & \textbf{for} $i=1$ \textbf{to} $K$ \textbf{do}\hspace{4pt}\algNote{($K$ can exceed the training depth)} & \\
        & \algN{9} & \hspace{8pt}$I_i^{\mathrm{in}}\gets\mathcal{Z}_{\gamma_i}(\hat I_{i-1})$; \ $p_i\gets G(\hat I_{i-1},I_i^{\mathrm{in}})$ & \algEnd{zoom, prompt} \\
        & \algN{10} & \hspace{8pt}$\hat I_i\gets D(F_{\theta}(I_i^{\mathrm{in}},p_i))$ & \algEnd{recursive SR} \\
    \end{tabular}
    \caption{\footnotesize \textbf{\method{}}~~\textit{Training and Inference}}
    \label{alg:oraclezoom}
\end{algorithm}

\vspace{3pt}

\noindent\textbf{\stepnum{1}~Supervision at Target-Available Scales.}
One shared adapter produces all predictions, so improving the deeper predictions must preserve the target-available ones. For $i \leq m$, we supervise predictions with the ground truth:
\begin{equation}
    \mathcal{L}_{\mathrm{sup}} = \frac{1}{m} \sum_{i=1}^{m} \operatorname{LPIPS} \left( \hat I_i, I_i^{\mathrm{gt}} \right).
    \label{eq:supervised_loss}
\end{equation}
This anchors the adapted SR model to observed image detail before direct supervision disappears beyond $s_m$.

\vspace{3pt}
\noindent\textbf{\stepnum{2}~EMA Consistency.}
To add a stable second target at $s_m$, an EMA copy $\bar{\theta}$ of the adapter receives $I_m^{\mathrm{gt}}$ and gives a training-only latent:
\begin{equation}
    z_m^{\mathrm{ema}} = \operatorname{sg}[F_{\bar{\theta}}(I_m^{\mathrm{gt}}, p_m)],\, \mathcal{L}_{\mathrm{ema}} = \mathrm{MSE}(z_m^\theta, z_m^{\mathrm{ema}}),
    \label{eq:ema_loss}
\end{equation}
where $\operatorname{sg}[\cdot]$ denotes stop-gradient.
After each optimization step, $\bar{\theta} \leftarrow \mu\bar{\theta} + (1-\mu)\theta$, with decay $\mu{=}0.95$.
This copy acts as a privileged self-teacher, an EMA variant of the teacher in OPSD~\cite{zhao2026selfdistilled} (details in the supplementary material).

\vspace{3pt}
\noindent\textbf{\stepnum{3}~Cross-Scale Ground-Truth Consistency.}
This objective is the core of \method{}: it keeps the last ground truth in use beyond $s_m$. For $j {>} m$, $I_j^{\mathrm{gt}}$ is unavailable, but the zoom path identifies the region of $I_m^{\mathrm{gt}}$ that $\hat I_j$ shows.
Let $r{=}s_j/s_m$. We define
\begin{equation}
    I_{m\rightarrow j}^{\mathrm{ref}} = \mathcal{A}_{m\rightarrow j} \left( I_m^{\mathrm{gt}} \right), \qquad \hat I_{j\rightarrow m} = \mathcal{P}_{j\rightarrow m} \left( \hat I_j \right),
    \label{eq:cross_scale_mapping}
\end{equation}
The alignment operator $\mathcal{A}_{m\rightarrow j}$ takes the center crop of side $H/r$ from $I_m^{\mathrm{gt}}$. This crop is the region that the zoom path magnifies into $\hat I_j$. The projection operator $\mathcal{P}_{j\rightarrow m}$ downsamples $\hat I_j$ by $r$ with bicubic interpolation to the same $H/r$ size.
Both images then show the same scene region at the resolution where ground truth exists, and we impose
\begin{equation}
    \mathcal{L}_{\mathrm{xscale}} = \frac{1}{K-m} \sum_{j=m+1}^{K} \operatorname{LPIPS} \left( \hat I_{j\rightarrow m}, I_{m\rightarrow j}^{\mathrm{ref}} \right).
    \label{eq:cross_scale_loss}
\end{equation}
The last target stays a visual reference beyond $s_m$: predictions there remain constrained by verifiable ground-truth information without requiring $I_j^{\mathrm{gt}}$.

\vspace{3pt}
\noindent\textbf{\stepnum{4}~Quality-Guided Detail Synthesis.}
Cross-scale projection cannot fully constrain high-frequency detail, since different fine-scale predictions may produce similar lower-resolution observations.
We therefore use a frozen no-reference quality model $\mathcal{Q}$:
\begin{equation}
    \mathcal{L}_{\mathrm{qual}} = - \frac{1}{K-m} \sum_{j=m+1}^{K} \mathcal{Q} \left( \hat I_j \right).
    \label{eq:quality_loss}
\end{equation}
This encourages perceptually detailed predictions where direct high-resolution supervision is unavailable.

\vspace{3pt}
\noindent\textbf{\stepnum{5}~Latent Prior from the Base Model.}
Optimizing a learned quality score can reward patterns that score well but are not supported by the input. This is a form of reward hacking.
We therefore keep each deeper latent close to the latent of the frozen base model $F_0$ (without the adapter) for the same input and prompt.
$F_0$ was trained on large paired data and has never seen the quality score. Its output is a conservative reference: less detailed, but free of score-driven artifacts.
With $z_j^0 {=} \operatorname{sg}[F_0(I_j^{\mathrm{in}},p_j)]$, we impose
\begin{equation}
    \mathcal{L}_{\mathrm{prior}} = \frac{1}{K-m} \sum_{j=m+1}^{K} \operatorname{MSE} \left( z_j^\theta, z_j^0 \right).
    \label{eq:prior_loss}
\end{equation}
While $\mathcal{L}_{\mathrm{qual}}$ encourages detail that $I_m^{\mathrm{gt}}$ cannot verify, $\mathcal{L}_{\mathrm{prior}}$ keeps that detail close to what $F_0$ would produce.

\begin{table*}[t]
\centering
\scriptsize
\setlength{\tabcolsep}{2.2pt}
\renewcommand{\arraystretch}{1.05}

\resizebox{\textwidth}{!}{%
\begin{tabular}{l cccccc cccc cccccc}
\toprule
\multirow{4}{*}{Method}
& \multicolumn{6}{c}{\textit{In-domain}}
& \multicolumn{10}{c}{\textit{Out-of-domain}}
\\
\cmidrule(lr){2-7} \cmidrule(l){8-17}
& \multicolumn{6}{c}{\textbf{4KLSDB}}
& \multicolumn{4}{c}{\textbf{DIV2K}}
& \multicolumn{6}{c}{\textbf{DIV8K}}
\\
\cmidrule(lr){2-7} \cmidrule(lr){8-11} \cmidrule(l){12-17}
& \multicolumn{4}{c}{No-reference quality}
& \multicolumn{2}{c}{GT fidelity}
& \multicolumn{4}{c}{No-reference quality}
& \multicolumn{4}{c}{No-reference quality}
& \multicolumn{2}{c}{GT fidelity}
\\
\cmidrule(lr){2-5} \cmidrule(lr){6-7} \cmidrule(lr){8-11} \cmidrule(lr){12-15} \cmidrule(l){16-17}
& NIQE$\downarrow$ & MUSIQ$\uparrow$ & MANIQA$\uparrow$ & CLIPIQA$\uparrow$ & LPIPS$\downarrow$ & DISTS$\downarrow$
& NIQE$\downarrow$ & MUSIQ$\uparrow$ & MANIQA$\uparrow$ & CLIPIQA$\uparrow$
& NIQE$\downarrow$ & MUSIQ$\uparrow$ & MANIQA$\uparrow$ & CLIPIQA$\uparrow$ & LPIPS$\downarrow$ & DISTS$\downarrow$
\\
\midrule
\rowcolor{gray!22}
\multicolumn{17}{l}{\textbf{$4\times$ magnification}} \\
HiT-SR~\cite{zhang2024hierarchical} & 8.76 & 40.11 & 0.396 & 0.396 & 0.408 & 0.232 & 7.89 & 39.21 & 0.403 & 0.374 & 8.05 & 37.06 & 0.395 & 0.366 & 0.448 & 0.265 \\
MambaIR~\cite{guo2024mambair} & 8.68 & 41.24 & 0.405 & 0.412 & 0.412 & 0.233 & 7.81 & 40.57 & 0.419 & 0.386 & 8.00 & 38.53 & 0.412 & 0.377 & 0.449 & 0.268 \\
SwinIR~\cite{liang2021swinir} & 6.02 & 51.31 & 0.500 & 0.465 & 0.280 & 0.184 & 5.66 & 51.00 & 0.510 & 0.458 & 5.85 & 49.42 & 0.512 & 0.447 & 0.320 & 0.210 \\
SeeSR~\cite{wu2024seesr} & \underline{5.44} & \underline{63.47} & \underline{0.569} & 0.615 & 0.260 & \underline{0.174} & 4.70 & 64.26 & 0.608 & 0.619 & \underline{4.66} & 63.94 & 0.612 & 0.610 & 0.263 & 0.174 \\
OSEDiff~\cite{wu2024one} & \textbf{5.00} & 60.31 & 0.559 & 0.608 & 0.335 & 0.215 & 4.93 & 59.54 & 0.585 & 0.638 & 4.88 & 60.59 & 0.596 & 0.639 & 0.380 & 0.243 \\
CoZ~\cite{kim2025chain} & 5.74 & \textbf{64.38} & \textbf{0.582} & \underline{0.623} & \underline{0.248} & 0.186 & \underline{4.65} & \textbf{66.94} & \underline{0.633} & \underline{0.699} & 4.67 & \underline{67.04} & \underline{0.638} & \underline{0.697} & \underline{0.238} & \underline{0.167} \\
\rowcolor{oursbg}
\textbf{\method{}} & 6.00 & 59.80 & 0.556 & \textbf{0.638} & \textbf{0.169} & \textbf{0.142} & \textbf{4.46} & \underline{66.47} & \textbf{0.638} & \textbf{0.735} & \textbf{4.48} & \textbf{67.31} & \textbf{0.648} & \textbf{0.736} & \textbf{0.223} & \textbf{0.161} \\
\midrule
\rowcolor{gray!22}
\multicolumn{17}{l}{\textbf{$16\times$ magnification}} \\
HiT-SR~\cite{zhang2024hierarchical} & 14.13 & 19.70 & 0.333 & 0.370 & -- & -- & 13.11 & 18.39 & 0.310 & 0.312 & 13.27 & 18.99 & 0.314 & 0.318 & -- & -- \\
MambaIR~\cite{guo2024mambair} & 14.42 & 20.40 & 0.333 & 0.390 & -- & -- & 13.28 & 19.23 & 0.312 & 0.329 & 13.42 & 19.78 & 0.315 & 0.334 & -- & -- \\
SwinIR~\cite{liang2021swinir} & 7.34 & 28.38 & 0.427 & 0.454 & -- & -- & 6.78 & 28.96 & 0.391 & 0.401 & 7.21 & 29.18 & 0.388 & 0.391 & -- & -- \\
SeeSR~\cite{wu2024seesr} & 7.81 & 47.83 & 0.507 & 0.552 & -- & -- & 6.50 & 51.78 & 0.523 & 0.540 & 6.49 & 52.96 & 0.528 & 0.535 & -- & -- \\
OSEDiff~\cite{wu2024one} & \textbf{6.10} & 51.92 & 0.531 & \underline{0.586} & -- & -- & \underline{5.96} & 53.01 & 0.545 & 0.605 & \textbf{5.79} & 53.48 & 0.550 & 0.611 & -- & -- \\
CoZ~\cite{kim2025chain} & 8.08 & \underline{52.26} & \underline{0.551} & 0.578 & -- & -- & 6.29 & \underline{58.49} & \underline{0.595} & \underline{0.652} & 6.32 & \underline{58.89} & \underline{0.602} & \underline{0.663} & -- & -- \\
\rowcolor{oursbg}
\textbf{\method{}} & \underline{7.16} & \textbf{54.80} & \textbf{0.556} & \textbf{0.658} & -- & -- & \textbf{5.82} & \textbf{61.64} & \textbf{0.599} & \textbf{0.737} & \underline{5.85} & \textbf{61.56} & \textbf{0.605} & \textbf{0.735} & -- & -- \\
\midrule
\rowcolor{gray!22}
\multicolumn{17}{l}{\textbf{$64\times$ magnification}} \\
HiT-SR~\cite{zhang2024hierarchical} & 16.69 & 23.76 & 0.388 & 0.455 & -- & -- & 16.52 & 21.82 & 0.367 & 0.408 & 16.55 & 22.03 & 0.370 & 0.408 & -- & -- \\
MambaIR~\cite{guo2024mambair} & 17.82 & 23.33 & 0.388 & 0.469 & -- & -- & 17.10 & 21.64 & 0.366 & 0.419 & 17.09 & 21.77 & 0.369 & 0.418 & -- & -- \\
SwinIR~\cite{liang2021swinir} & 9.02 & 26.52 & 0.489 & 0.477 & -- & -- & 8.09 & 23.20 & 0.478 & 0.465 & 8.62 & 22.93 & 0.474 & 0.460 & -- & -- \\
SeeSR~\cite{wu2024seesr} & 9.83 & 37.31 & 0.495 & 0.497 & -- & -- & 8.98 & 42.00 & 0.502 & 0.517 & 8.88 & 43.76 & 0.509 & 0.524 & -- & -- \\
OSEDiff~\cite{wu2024one} & \textbf{7.20} & \underline{46.22} & 0.531 & 0.533 & -- & -- & \underline{7.30} & 47.71 & 0.536 & 0.576 & \textbf{7.04} & 47.66 & 0.537 & 0.580 & -- & -- \\
CoZ~\cite{kim2025chain} & 9.54 & 45.70 & \underline{0.555} & \underline{0.544} & -- & -- & 7.86 & \underline{52.06} & \underline{0.581} & \underline{0.623} & 7.78 & \underline{51.99} & \underline{0.584} & \underline{0.630} & -- & -- \\
\rowcolor{oursbg}
\textbf{\method{}} & \underline{8.47} & \textbf{50.80} & \textbf{0.579} & \textbf{0.651} & -- & -- & \textbf{7.12} & \textbf{56.02} & \textbf{0.597} & \textbf{0.727} & \underline{7.23} & \textbf{55.61} & \textbf{0.597} & \textbf{0.726} & -- & -- \\
\midrule
\rowcolor{gray!22}
\multicolumn{17}{l}{\textbf{$256\times$ magnification}} \\
HiT-SR~\cite{zhang2024hierarchical} & 17.81 & 26.23 & 0.428 & 0.494 & -- & -- & 17.64 & 26.32 & 0.421 & 0.480 & 17.52 & 26.34 & 0.419 & 0.487 & -- & -- \\
MambaIR~\cite{guo2024mambair} & 18.97 & 26.04 & 0.426 & 0.509 & -- & -- & 18.47 & 26.23 & 0.418 & 0.493 & 18.43 & 26.39 & 0.416 & 0.497 & -- & -- \\
SwinIR~\cite{liang2021swinir} & 10.77 & 28.54 & 0.495 & 0.467 & -- & -- & 9.96 & 27.35 & 0.504 & 0.466 & 10.01 & 27.34 & 0.504 & 0.473 & -- & -- \\
SeeSR~\cite{wu2024seesr} & 11.56 & 33.66 & 0.503 & 0.496 & -- & -- & 11.16 & 36.26 & 0.502 & 0.497 & 10.79 & 37.64 & 0.505 & 0.510 & -- & -- \\
OSEDiff~\cite{wu2024one} & \textbf{8.18} & 42.57 & 0.525 & 0.502 & -- & -- & \underline{8.53} & 43.71 & 0.528 & 0.543 & \underline{8.25} & 43.98 & 0.528 & 0.551 & -- & -- \\
CoZ~\cite{kim2025chain} & 10.41 & \underline{44.13} & \underline{0.555} & \underline{0.538} & -- & -- & 9.24 & \underline{48.26} & \underline{0.576} & \underline{0.599} & 8.83 & \underline{48.57} & \underline{0.577} & \underline{0.608} & -- & -- \\
\rowcolor{oursbg}
\textbf{\method{}} & \underline{9.31} & \textbf{49.31} & \textbf{0.589} & \textbf{0.664} & -- & -- & \textbf{8.30} & \textbf{51.83} & \textbf{0.595} & \textbf{0.703} & \textbf{8.07} & \textbf{51.85} & \textbf{0.593} & \textbf{0.710} & -- & -- \\
\bottomrule
\end{tabular}%
}

\caption{
    \textbf{Per-dataset results across recursion depth.}
    LPIPS and DISTS need ground truth. Here, only 4KLSDB and DIV8K have it, and only at $4\times$, so deeper zooms show ``--''.
    The supplementary material gives the other four test sets.
    \textbf{Bold}: best. \underline{Underline}: second-best.
}
\label{tab:main}
\vspace{-3pt}
\end{table*}

\subsection{Overall Objective}
\label{sec:overall_objective}

We minimize the weighted sum of the five terms:
\begin{equation}
    \begin{aligned}
        \mathcal{L} ={}
        & \mathcal{L}_{\mathrm{sup}} + \lambda_{\mathrm{xscale}} \mathcal{L}_{\mathrm{xscale}} + \lambda_{\mathrm{qual}} \mathcal{L}_{\mathrm{qual}} \\
        & + \lambda_{\mathrm{prior}} \mathcal{L}_{\mathrm{prior}} + \lambda_{\mathrm{ema}} \mathcal{L}_{\mathrm{ema}}.
    \end{aligned}
    \label{eq:total_objective}
\end{equation}
We optimize \cref{eq:total_objective} with fixed weights.
$\mathcal{L}_{\mathrm{sup}}$, $\mathcal{L}_{\mathrm{xscale}}$, and $\mathcal{L}_{\mathrm{prior}}$ act as soft constraints while $\mathcal{L}_{\mathrm{qual}}$ improves the deeper predictions.

\subsection{Training and Inference}
\label{sec:training_inference}

\Cref{alg:oraclezoom} summarizes training and inference for \method{}.
During training, the adapter parameters $\theta$ are shared across scales. The SR backbone, VAE decoder $D$, VLM prompter $G$, quality model $\mathcal{Q}$, and base model $F_0$ remain frozen.
Only $\theta$ is optimized by gradients; $\bar{\theta}$ is updated by EMA.
We backpropagate through successive predictions, allowing deeper-scale objectives to also update earlier steps.
Ground-truth targets, aligned references, $\mathcal{Q}$, $F_0$, and the EMA branch are used only for training.

At inference, the learned adapter is used with the frozen SR backbone, decoder, and VLM prompter over scales $s_1,\ldots,s_K$.
\method{} needs no ground truth or training-only branches at inference.

\section{Experiments}
\label{sec:exp}
We evaluate \method{} on seven benchmarks across four magnifications and ask whether it improves perceptual quality at every zoom depth, preserves the evidence in the last ground truth, and limits hallucination. Ablations then isolate the role of each objective.

\begin{figure*}[t]
    \centering
    \includegraphics[width=\linewidth]{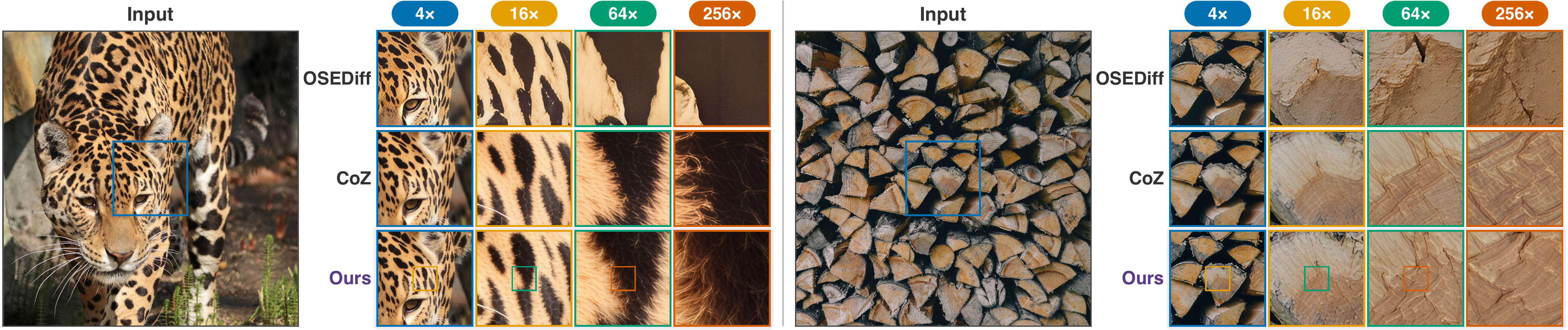}
    \caption{
    \textbf{Qualitative comparison from $4\times$ to $256\times$.}
    Colored boxes mark the region magnified at the next step.
    \method{} keeps dense fur and wood structure through $256\times$, while OSEDiff and CoZ flatten or distort them.
}
    \label{fig:qual}
    \vspace{-10pt}
\end{figure*}

\begin{figure*}[t]
    \centering
    \includegraphics[width=\linewidth]{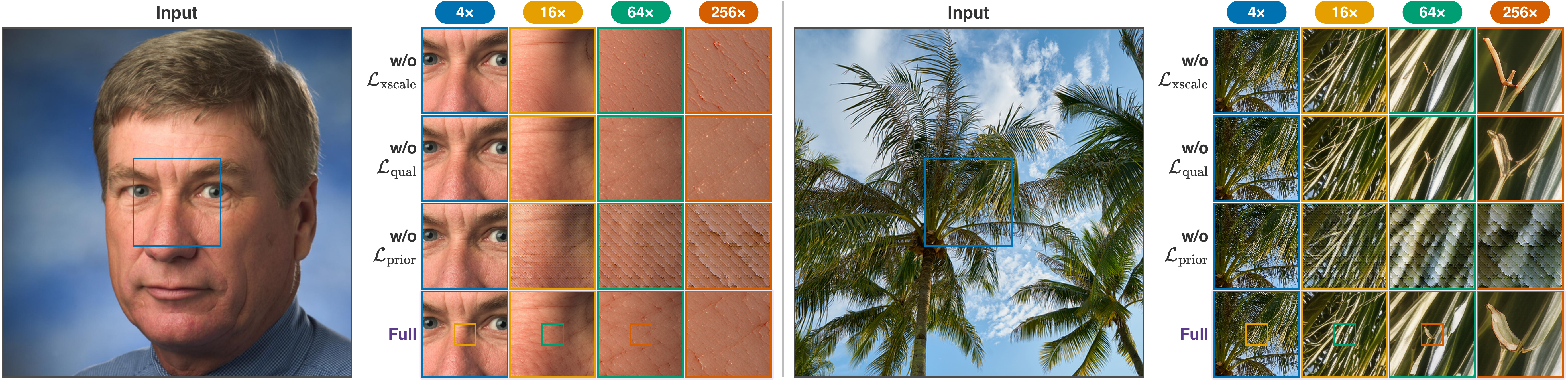}
    \caption{
    \textbf{Visual ablation of the deep-scale objectives.}
    Colored boxes as in \cref{fig:qual}.
    Without $\mathcal{L}_{\mathrm{xscale}}$, the zoom drifts from the real scene. Without $\mathcal{L}_{\mathrm{qual}}$, the skin turns smooth. Without $\mathcal{L}_{\mathrm{prior}}$, a repetitive fish-scale pattern appears from $16\times$ on.
}
    \label{fig:objective-qual}
    \vspace{-5pt}
\end{figure*}

\subsection{Experimental Setup}
\label{sec:setup}

\noindent\textbf{Datasets.}
For training \method{}, we sample a \trainsize{}-image training set from $129{,}484$ candidates in the 4KLSDB~\cite{zhu20264klsdb} training split.
We keep images with a short side of at least $2048$ pixels and discard the lowest-quality $10\%$.
To improve diversity, we balance content groups derived from captions.
We keep only images that both DFN5B and SigLIP2 classify as photographic.
Using image similarity and CLIP verification, we also remove near-duplicates of any evaluation image.
We evaluate on 4KLSDB~\cite{zhu20264klsdb}, DIV2K~\cite{agustsson2017ntire}, DIV8K~\cite{gu2019div8k}, DRealSR~\cite{wei2020component}, RealSR~\cite{cai2019toward}, FFHQ~\cite{karras2019style}, and Flickr2K~\cite{lim2017edsr}, with $200$, $800$, $1238$, $93$, $100$, $200$, and $200$ images, respectively.
Each input is the {\footnotesize $512{\times}512$} center crop after a LANCZOS resize of the short side to $512$. On DRealSR and RealSR, we use only the high-resolution images.
Following CoZ~\cite{kim2025chain}, every method processes the same {\footnotesize $512{\times}512$} center crop through four {\footnotesize $4\times$} SR steps, producing {\footnotesize $4\times$, $16\times$, $64\times$}, and {\footnotesize $256\times$} outputs.

\vspace{3pt}
\noindent\textbf{Evaluation Metrics.}
For no-reference quality at all scales, including those without ground truth ($16\times$, $64\times$, $256\times$), we use NIQE~\cite{mittal2013making} ($\downarrow$), MUSIQ~\cite{ke2021musiq} ($\uparrow$), MANIQA~\cite{yang2022maniqa} ($\uparrow$), and CLIPIQA~\cite{wang2023clipiqa} ($\uparrow$).
TOPIQ-NR~\cite{chen2024topiq} supplies $\mathcal{L}_{\mathrm{qual}}$, so the primary evaluation uses the four metrics above.
At $4\times$, where high-resolution ground truth is available, we measure fidelity on 4KLSDB, DIV8K, DRealSR, and RealSR using LPIPS~\cite{zhang2018unreasonable} ($\downarrow$) and DISTS~\cite{ding2022image} ($\downarrow$). DINOv2 cosine similarity~\cite{oquab2024dinov2} ($\uparrow$) is in the supplementary material. 
At $16\times$, we project each prediction back to the supervision boundary ($4\times$).
We compare this projected prediction with the aligned reference using projected DISTS (P-DISTS).
Aggregate GT-fidelity numbers count all $1{,}631$ images of these four sets equally.
We measure hallucination with two \emph{ground-truth-anchored judges} from different vendors, Gemini 4 Argon~\cite{google2026gemini4argon} and GPT-6.1 Sol~\cite{openai2026gpt61sol}.
We use $120$ randomly sampled, region-aligned items per scale, $20$ from each of the six test sets other than DIV2K.
In each item, a judge sees the native ground-truth window that the zoom step magnifies (the \emph{anchor}) and two labeled candidates (a method and CoZ).
Each pair is judged in both A/B orders.
A candidate \emph{hallucinates} when the judge flags colors, large structures, edges, objects, or patterns that contradict the anchor.
Fine detail counts as faithful if the anchor is too blurry to show it.
The hallucination rate is the fraction of judgments in which a method is flagged, pooled over both judges ($480$ judgments per scale).
The win rate ignores ties.
When both judges pick a winner, they pick the same one in $75\%$ of cases.
A validation with known answers (supplementary material) shows that both judges separate real photographs from model outputs when the window holds about $256$ native pixels or more, but not at about $64$.
We therefore treat results with small windows as weak evidence.

\vspace{3pt}
\noindent\textbf{Implementation Details.}
For a controlled comparison, we apply \method{} to the frozen models of CoZ: one-step OSEDiff~\cite{wu2024one} on SD3-medium, the GRPO-tuned Qwen2.5-VL-3B-Instruct prompter, and the VAE decoder.
We adapt the SD3 transformer with a rank-$16$ LoRA~\cite{hu2022lora}, which has $7.1$M trainable parameters.
At inference, the adapter adds $0.019$\,s per recursive step to CoZ ($0.001$\,s after merging the LoRA weights; supplementary material).
We train the shared adapter across the ${4\times} {\rightarrow} {16\times}$ recursive chain ($m{=}1$, $K{=}2$) and backpropagate through the $4\times$ prediction.
Inference uses $K{=}4$.
We set $\lambda_{\mathrm{xscale}}{=}1.0$, $\lambda_{\mathrm{qual}}{=}0.4$, $\lambda_{\mathrm{prior}}{=}8.0$, and $\lambda_{\mathrm{ema}}{=}0.1$, with an EMA decay of $0.95$.
We optimize using AdamW with a learning rate of $5{\times}10^{-5}$, weight decay of $10^{-2}$, a $500$-step warmup, and an effective batch size of $4$.
The final checkpoints are selected by early stopping on the loss of \cref{eq:total_objective} over a held-out $2{,}000$-image validation split, after about $9.3$k steps. The loss weights were chosen on the same split.
Full implementation details are in the supplementary material. We will release the code and trained adapter.

\noindent\textbf{Baselines.}
We compare with three feed-forward models (SwinIR~\cite{liang2021swinir}, HiT-SR~\cite{zhang2024hierarchical}, and MambaIR~\cite{guo2024mambair}), two diffusion models (SeeSR~\cite{wu2024seesr} and OSEDiff~\cite{wu2024one}), and CoZ~\cite{kim2025chain}. The first five use the public 4KLSDB-fine-tuned checkpoints~\cite{zhu20264klsdb}, and OSEDiff keeps its SD2.1 backbone.
Every baseline uses the same CoZ recursion, with matched inputs, zoom paths, crops, prompts, and metrics.

\begin{table}[t]
\centering
\scriptsize
\setlength{\tabcolsep}{2.6pt}
\renewcommand{\arraystretch}{1.08}

\resizebox{\columnwidth}{!}{%
\begin{tabular}{l cccc cc}
\toprule
\multirow{2}{*}{Method}
& \multicolumn{4}{c}{No-reference quality}
& \multicolumn{2}{c}{GT fidelity @$4\times$}
\\
\cmidrule(lr){2-5} \cmidrule(l){6-7}
& NIQE$\downarrow$ & MUSIQ$\uparrow$ & MANIQA$\uparrow$ & CLIPIQA$\uparrow$ & LPIPS$\downarrow$ & DISTS$\downarrow$
\\
\midrule
HiT-SR~\cite{zhang2024hierarchical} & 14.22 & 27.06 & 0.379 & 0.414 & 0.420 & 0.253 \\
MambaIR~\cite{guo2024mambair} & 14.69 & 27.34 & 0.382 & 0.428 & 0.421 & 0.255 \\
SwinIR~\cite{liang2021swinir} & 7.88 & 34.02 & 0.477 & 0.456 & 0.295 & 0.200 \\
SeeSR~\cite{wu2024seesr} & 8.35 & 48.46 & 0.532 & 0.546 & 0.252 & 0.171 \\
OSEDiff~\cite{wu2024one} & \textbf{6.70} & 51.38 & 0.548 & 0.581 & 0.366 & 0.237 \\
CoZ~\cite{kim2025chain} & 7.42 & \underline{55.23} & \underline{0.587} & \underline{0.621} & \underline{0.233} & \underline{0.169} \\
\rowcolor{oursbg}
\textbf{\method{}} & \underline{6.90} & \textbf{57.80} & \textbf{0.598} & \textbf{0.713} & \textbf{0.214} & \textbf{0.160} \\
\bottomrule
\end{tabular}%
}

\caption{
    \textbf{Aggregate results.}
    We average no-reference metrics over the seven test sets and four scales. GT fidelity counts all $1{,}631$ images of the four sets with $4\times$ targets equally.
    \textbf{Bold}: best. \underline{Underline}: second-best.
}
\label{tab:agg7}
\vspace{-10pt}
\end{table}

\subsection{Results}
\label{sec:main}

\Cref{tab:main} reports no-reference quality at all four zoom scales and GT fidelity at $4\times$, and \cref{tab:agg7} gives the averages.

\vspace{3pt}
\noindent\textbf{Quality Across Recursion Depth.}
\method{} achieves the highest CLIPIQA at every zoom depth on all seven test sets.
From $16\times$ onward, \method{} also ranks first in MUSIQ and MANIQA on all three sets in \cref{tab:main}.
At $4\times$, it achieves the best LPIPS and DISTS on 4KLSDB and on the out-of-domain DIV8K. On DIV8K, it also leads on all four no-reference metrics.
The no-reference gains generalize beyond the training domain. At $4\times$, \method{} leads CoZ in CLIPIQA by $0.036$ on DIV2K and $0.039$ on DIV8K, against $0.015$ on the in-domain set. From $16\times$ on, its lead is at least $0.072$ on every out-of-domain set.
Averaged over all seven test sets (\cref{fig:teaser}, top right), \method{} stays above $0.70$ CLIPIQA throughout the recursion and obtains $0.706$ at $256\times$, compared with $0.579$ for CoZ, $0.532$ for OSEDiff, and $0.463$ for SwinIR.
In \cref{tab:agg7}, \method{} leads on MUSIQ, MANIQA, CLIPIQA, LPIPS, and DISTS, and is second on NIQE.

\begin{figure}[t]
    \centering
    \includegraphics[width=\linewidth]{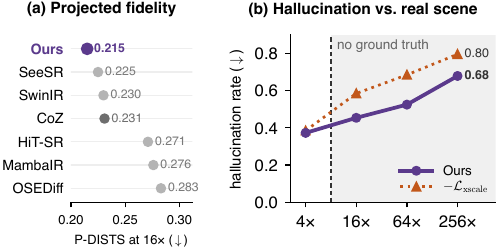}
    \caption{
    \textbf{Fidelity and hallucination beyond the supervision boundary.}
    \textbf{(a)} P-DISTS at $16\times$, mean over the $1{,}631$ images of the four sets with $4\times$ targets.
    \textbf{(b)} Hallucination rate pooled over two judges ($480$ judgments per scale), with and without $\mathcal{L}_{\mathrm{xscale}}$.
}
    \label{fig:results_depth}
    \vspace{-10pt}
\end{figure}

\vspace{3pt}
\noindent\textbf{Fidelity Beyond the Supervision Boundary.}
At $16\times$, the region remains visible in the $4\times$ ground truth, so we compare each projected prediction with the aligned reference.
In \cref{fig:results_depth}(a), \method{} achieves the lowest P-DISTS ($0.215$), against $0.225$ for SeeSR, $0.230$ for SwinIR, and $0.231$ for CoZ.
It best preserves the last ground-truth evidence among these methods (see supplementary material).

\vspace{3pt}
\noindent\textbf{Hallucination Against the Real Scene.}
\method{} raises no-reference quality at a hallucination rate equal to or below that of CoZ up to $16\times$.
At $4\times$, where ground truth is available, \method{} hallucinates less than CoZ ($0.37$ vs.\ $0.42$) and wins $69\%$ of decided comparisons.
At $16\times$ and $64\times$, the two methods are comparable ($0.45$ vs.\ $0.45$ and $0.53$ vs.\ $0.53$).
At $256\times$, \method{} is flagged more often ($0.68$ vs.\ $0.60$). However, the anchor holds very few native pixels.
Faces cause the difference at deep zooms (supplementary material).
On the five non-face sets, \method{} is flagged less often than CoZ up to $64\times$ and equally often at $256\times$ ($0.44$ vs.\ $0.50$, $0.47$ vs.\ $0.50$, $0.48$ vs.\ $0.60$, and $0.66$ vs.\ $0.66$).
\Cref{fig:results_depth}(b) shows what the cross-scale loss contributes: without $\mathcal{L}_{\mathrm{xscale}}$, the hallucination rate rises at every deeper zoom. At $16\times$, it rises from $0.45$ to $0.59$.

\vspace{3pt}
\noindent\textbf{Qualitative Results.}
\Cref{fig:qual} shows two examples through the ${4\times \to 16\times \to 64\times \to 256\times}$ trajectory. 
On the jaguar, OSEDiff flattens the fur into blank regions and CoZ turns it into stray strands.
On the woodpile, OSEDiff turns the wood into cracked mud and CoZ adds marble-like bands.
\method{} keeps dense fur and wood through $256\times$.

\begin{table}[t]
    \centering
    \scriptsize
    \setlength{\tabcolsep}{1.9pt}
    \renewcommand{\arraystretch}{1.08}
    \resizebox{\columnwidth}{!}{%
        \begin{tabular}{lccccc}
            \toprule
            \textbf{Variant} & \textbf{LPIPS$_{4\times}\downarrow$} & \textbf{P-DISTS$_{16\times}\downarrow$} & \textbf{CLIPIQA$_{16\times}\uparrow$} & \textbf{Hal.$^{\mathrm{GT}}_{16\times}\downarrow$} & \textbf{Hal.$^{\mathrm{GT}}_{64\times}\downarrow$} \\
            \midrule
            \rowcolor{oursbg} \textbf{Full \method{}} & $0.214$ & $0.215$ & $0.710$ & $0.45$ & $0.53$ \\
            $-\mathcal{L}_{\mathrm{sup}}$ & $0.228$ & $0.221$ & $0.719$ & $0.63$ & $0.68$ \\
            $-\mathcal{L}_{\mathrm{xscale}}$ & $0.213$ & $0.229$ & $0.727$ & $0.59$ & $0.69$ \\
            $-\mathcal{L}_{\mathrm{qual}}$ & $0.215$ & $\mathbf{0.210}$ & $0.594$ & $\mathbf{0.34}$ & $\mathbf{0.41}$ \\
            $-\mathcal{L}_{\mathrm{prior}}$ & $\mathbf{0.203}$ & $0.331$ & $\mathbf{0.794}$ & $1.00$ & $0.99$ \\
            $-\mathcal{L}_{\mathrm{ema}}$ & $0.217$ & $0.217$ & $0.702$ & $0.44$ & $0.53$ \\
            \midrule
            Fine-tuning only & $0.200$ & $0.198$ & $0.453$ & $0.27$ & $0.25$ \\
            Detached recursion & $0.215$ & $0.212$ & $0.709$ & $0.44$ & $0.55$ \\
            \bottomrule
        \end{tabular}%
    }
    \caption{
    \textbf{Ablation of the training objectives.}
    Top: each row removes one term from \cref{eq:total_objective} and keeps the other weights fixed.
    Bottom: \emph{fine-tuning only} trains with $\mathcal{L}_{\mathrm{sup}}$ at $4\times$ alone, and \emph{detached recursion} stops gradients between zoom steps.
    Hal.$^{\mathrm{GT}}$: hallucination rate of the ground-truth-anchored judges.
    LPIPS and P-DISTS count all $1{,}631$ images of the four sets with $4\times$ targets equally. CLIPIQA and Hal.$^{\mathrm{GT}}$ use the six test sets other than DIV2K.
    Bold: best leave-one-out.
}
    \label{tab:analysis}
\end{table}

\subsection{Ablation Studies}
\label{sec:analysis}

\begin{figure}[t]
    \centering
    \includegraphics[width=\linewidth]{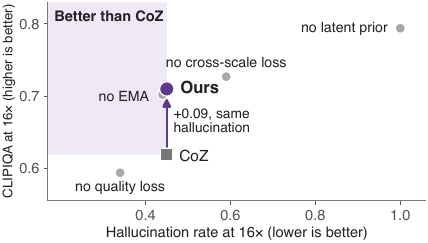}
    \caption{
\textbf{Quality against hallucination at $16\times$.}
    Each point is one variant from \cref{tab:analysis}, plus CoZ on the same six test sets.
    At the hallucination rate of CoZ, \method{} has higher CLIPIQA.
}
    \label{fig:tradeoff}
    \vspace{-10pt}
\end{figure}

\noindent\textbf{Training Objectives.}
\label{sec:ablations}
Each objective has a distinct role (\cref{tab:analysis}).
Removing $\mathcal{L}_{\mathrm{sup}}$ increases $4\times$ LPIPS from $0.214$ to $0.228$.
Removing $\mathcal{L}_{\mathrm{xscale}}$ increases P-DISTS from $0.215$ to $0.229$, and the hallucination rate (Hal.$^{\mathrm{GT}}$) grows with depth: $0.59$ at $16\times$ and $0.69$ at $64\times$, against $0.45$ and $0.53$ for the full model.
The aligned reference keeps deeper zooms tied to the real scene.
Without $\mathcal{L}_{\mathrm{qual}}$, hallucination is lowest among the leave-one-out variants, but CLIPIQA falls from $0.710$ to $0.594$, the expected trade-off between detail and faithfulness.
Removing $\mathcal{L}_{\mathrm{prior}}$ instead increases CLIPIQA to $0.794$, but P-DISTS degrades to $0.331$ and Hal.$^{\mathrm{GT}}$ reaches $1.00$ at $16\times$ and $0.99$ at $64\times$.
$\mathcal{Q}$ alone is not a reliable guard against artifacts, so we pair it with $\mathcal{L}_{\mathrm{prior}}$.
Removing $\mathcal{L}_{\mathrm{ema}}$ changes every metric only slightly, so this term is minor.
The supplementary material extends this to $64\times$ and $256\times$.
At $16\times$, every variant with higher CLIPIQA than the full objective also hallucinates more.
\Cref{fig:tradeoff} shows this trade-off: \method{} matches the hallucination rate of CoZ at higher quality.

\vspace{3pt}
\noindent\textbf{Training Scheme and Text Prompt.}
Two further ablations in \cref{tab:analysis} test where the gain comes from.
One ablation fine-tunes the adapter with $\mathcal{L}_{\mathrm{sup}}$ at $4\times$ alone, with no deep-scale loss.
It has better $4\times$ fidelity (LPIPS $0.200$ vs.\ $0.214$), but CLIPIQA at $16\times$ drops from $0.710$ to $0.453$. The deep-scale losses trade a small $4\times$ fidelity cost for a large quality gain.
Its deep zooms are smooth and add little detail, so they contradict the scene less often.
Stopping gradients between zoom steps changes quality and hallucination little.
The prompt matters mainly at $256\times$: with an empty prompt, \method{} drops from $0.706$ to $0.673$ CLIPIQA there.
This is still above $0.579$ for CoZ with its full prompter (supplementary material).

\vspace{3pt}
\noindent\textbf{Visual Ablation.}
\Cref{fig:objective-qual} shows the effect of each deep-scale objective. 
Removing $\mathcal{L}_{\mathrm{xscale}}$ lets deeper zooms drift from the real scene, changing the color and shape of fine structures. 
Removing $\mathcal{L}_{\mathrm{qual}}$ produces smooth predictions with limited fine-scale structure.
Removing $\mathcal{L}_{\mathrm{prior}}$ instead introduces repetitive patterns that score highly on the quality objective but contradict the preceding zoom. 
This matches \cref{tab:analysis}, where removing $\mathcal{L}_{\mathrm{prior}}$ gives the highest quality score but worse P-DISTS and more hallucination.

\section{Conclusion}
\label{sec:conclusion}
Recursive SR loses ground truth after the earliest zooms, but the last ground-truth image still contains the region that every deeper zoom magnifies.
\method{} turns this region into a training signal: it trains on its own recursive trajectory and projects each deeper prediction back to its aligned reference.
A no-reference quality objective guides the remaining detail, and a latent prior limits drift from the pretrained model.
Across seven datasets, \method{} achieves the highest mean CLIPIQA, MUSIQ, and MANIQA among the compared methods.
Pooled over two judges, it adds this detail at the same hallucination rate as CoZ at $16\times$ and a lower rate at $4\times$, and its cross-scale loss lowers hallucination at $16\times$.

\noindent\textbf{Limitations.}
Beyond $4\times$, P-DISTS and the judges measure consistency with the evidence that remains observable, and our evaluation simulates zoom with center crops.
Fidelity results on the real-camera sets DRealSR and RealSR are mixed.
Faces remain the main failure case: \method{} hallucinates more than CoZ on FFHQ at $64\times$ and $256\times$.
The supplementary material discusses these limits.

\section*{LLM Usage Statement}
The authors take full responsibility for all submitted content, regardless of how LLMs or other AI tools were used.
We gave the full manuscript or major parts of it to ChatGPT (OpenAI) to get reviewer-style feedback.
LLM tools generated parts of the text (language editing, restructuring, and drafting) and parts of the code.
The authors reviewed, edited, or approved all of this output.
LLM tools helped edit and refine the authors' own related-work section, citations, and figures.
No LLM or AI tool was used for experiment design, execution, or analysis.
None was used for data creation or annotation.
None was used for research ideas, hypotheses, claims, or conclusions.

{
    \small
    \bibliographystyle{ieeenat_fullname}
    \bibliography{main,appendix}
}

\clearpage
\appendix
\setcounter{table}{0}
\setcounter{figure}{0}
\renewcommand{\thetable}{S\arabic{table}}
\renewcommand{\thefigure}{S\arabic{figure}}
\renewcommand{\theHtable}{S\arabic{table}}
\renewcommand{\theHfigure}{S\arabic{figure}}
\renewcommand{\topfraction}{0.95}
\renewcommand{\dbltopfraction}{0.95}
\renewcommand{\bottomfraction}{0.9}
\renewcommand{\textfraction}{0.05}
\renewcommand{\floatpagefraction}{0.8}
\renewcommand{\dblfloatpagefraction}{0.8}
\setcounter{topnumber}{3}
\setcounter{dbltopnumber}{3}
\setcounter{totalnumber}{5}
\makeatletter
\setlength{\@fptop}{0pt}
\setlength{\@fpsep}{12pt}
\setlength{\@fpbot}{0pt plus 1fil}
\setlength{\@dblfptop}{0pt}
\setlength{\@dblfpsep}{24pt}
\setlength{\@dblfpbot}{0pt plus 1fil}
\makeatother

\twocolumn[{\centering\Large\bfseries \method{}: Reference-Constrained Recursive Image Super-Resolution Beyond Ground Truth: \textit{\textcolor{blue}{Supplementary Material}}\par\vspace{2em}}]

\noindent\textbf{Supplementary Overview.}
This supplementary material gives more method details and analysis (\cref{app:method}), implementation and evaluation details (\cref{app:implementation}), more quantitative and qualitative results and analysis (\cref{app:additional_results}), limitations (\cref{app:limitations}), and the judge prompts (\cref{app:prompts}).

\section{Additional Method Details and Analysis}
\label{app:method}

This section provides additional details and analysis for \method{}.
We use the notation of the Method section of the main paper.
We first show how the last available ground truth remains useful at deeper recursive scales and why it cannot constrain fine detail.
We then describe the EMA term and relate \method{} to prior self-distillation approaches.

\subsection{Using the Last Ground Truth at Deeper Zooms}
\label{app:cross_scale_reference}

Recall that $I_m^{\mathrm{gt}}$ is the last available ground-truth target, while no target $I_j^{\mathrm{gt}}$ exists for $j {>} m$.
Although $I_m^{\mathrm{gt}}$ exists only at scale $s_m$, it still contains the image region that every later deterministic zoom selects.
The available target can therefore remain useful after direct supervision ends.
For a deeper scale $s_j$, define its magnification relative to $s_m$ as
\begin{equation}
r
=
\frac{s_j}{s_m}
=
\prod_{k=m+1}^{j}
\gamma_k.
\label{eq:appendix_cumulative_zoom}
\end{equation}
The zoom path from $s_m$ to $s_j$ selects the center region of $I_m^{\mathrm{gt}}$ with side $H/r$.
We write this unresized crop as
\begin{equation}
\mathcal{A}_{m\rightarrow j}(X)
=
\operatorname{CenterCrop}_{H/r}(X),
\label{eq:appendix_alignment_operator}
\end{equation}
which uses the same deterministic spatial selection as recursive SR.
Applying $\mathcal{A}_{m\rightarrow j}$ to the last available target gives
\begin{equation}
I_{m\rightarrow j}^{\mathrm{ref}}
=
\mathcal{A}_{m\rightarrow j}
\left(
I_m^{\mathrm{gt}}
\right).
\label{eq:appendix_reference}
\end{equation}

The prediction $\hat I_j$ shows the same field of view as $I_{m\rightarrow j}^{\mathrm{ref}}$, but at scale $s_j$.
We downsample $\hat I_j$ by a factor of $r$, to side $H/r$.
This step projects $\hat I_j$ back to the resolution of $s_m$:
\begin{equation}
\begin{aligned}
\mathcal{P}_{j\rightarrow m}
&=
\operatorname{Downsample}_{j\rightarrow m},
\\
\hat I_{j\rightarrow m}
&=
\mathcal{P}_{j\rightarrow m}
\left(
\hat I_j
\right).
\end{aligned}
\label{eq:appendix_projection}
\end{equation}
Consequently, $I_{m\rightarrow j}^{\mathrm{ref}}$ and $\hat I_{j\rightarrow m}$ show the same spatial region at the same resolution: the resolution of $s_m$, where ground truth exists.

\subsection{Limits of Cross-Scale Consistency}
\label{app:cross_scale_ambiguity}

Cross-scale consistency preserves the information that can still be verified from $I_m^{\mathrm{gt}}$, but it cannot determine all fine-scale detail at $s_j$.
Multiple high-resolution images can produce the same image after projection.
Consider the projection $\mathcal{P}_{j\rightarrow m}:\mathcal{X}_j\rightarrow\mathcal{X}_m$.
Here, $\mathcal{X}_j$ is the set of $H{\times}H$ images at scale $s_j$, and $\mathcal{X}_m$ is the set of crops of side $H/r$ at the resolution of $s_m$.
Since $s_j {>} s_m$, the projection removes fine-scale information.
For fixed image sizes, bicubic downsampling is a linear map.
This map has a nontrivial null space, so some nonzero changes map to zero.
For any fine-scale change $\delta$ with $\mathcal{P}_{j\rightarrow m}(\delta){=}0$,
\begin{equation}
\mathcal{P}_{j\rightarrow m}(X+\delta)
=
\mathcal{P}_{j\rightarrow m}(X),
\label{eq:appendix_projection_ambiguity}
\end{equation}
so any loss after projection scores $X$ and $X+\delta$ equally.

Minimizing $\mathcal{L}_{\mathrm{xscale}}$ therefore constrains only what survives projection.
$\mathcal{L}_{\mathrm{qual}}$ scores the full prediction $\hat I_j$, so it also guides the detail that projection removes.
$\mathcal{L}_{\mathrm{prior}}$ keeps that detail close to $F_0$.

\subsection{EMA Consistency at the Boundary}
\label{app:ema}

The transition from $s_m$ to $s_{m+1}$ is where direct supervision ends: $s_m$ still has exact ground truth, while $s_{m+1}$ does not.
We therefore apply EMA consistency at this boundary.

The EMA parameters are updated as
\begin{equation}
\bar{\theta}_{t}
=
\mu\bar{\theta}_{t-1}
+
(1-\mu)\theta_t,
\label{eq:appendix_ema_update}
\end{equation}
and are not updated by gradients.
Since $\bar{\theta}$ averages the model parameters over training, it changes more slowly than $\theta$.
We use it to form a stable latent reference
\begin{equation}
z_m^{\mathrm{ema}}
=
\operatorname{sg}
\left[
F_{\bar{\theta}}
\left(
I_m^{\mathrm{gt}},
p_m
\right)
\right],
\label{eq:appendix_ema_target}
\end{equation}
and encourage the current representation to remain close to this reference using
\begin{equation}
\mathcal{L}_{\mathrm{ema}}
=
\operatorname{MSE}
\left(
z_m^\theta,
z_m^{\mathrm{ema}}
\right).
\label{eq:appendix_ema_loss}
\end{equation}

The EMA term adds a slowly varying target where direct supervision ends.
It complements $\mathcal{L}_{\mathrm{sup}}$.
The loss $\mathcal{L}_{\mathrm{sup}}$ matches the prediction to the available ground truth.
The loss $\mathcal{L}_{\mathrm{ema}}$ adds a second target in latent space at $s_m$.
It is a minor term with a small weight ($\lambda_{\mathrm{ema}}{=}0.1$) that slightly improves fidelity.

\subsection{Putting the Objectives Together}
\label{app:objective_summary}

The five objectives play different roles depending on whether ground truth is available.
For $i \leq m$, $\mathcal{L}_{\mathrm{sup}}$ directly matches the prediction to the observed target.
For $j {>} m$, $\mathcal{L}_{\mathrm{xscale}}$ carries forward the information from $I_m^{\mathrm{gt}}$ that can still be verified along the recursive zoom path.
This evidence does not determine all fine-scale detail.
Therefore, $\mathcal{L}_{\mathrm{qual}}$ guides the remaining detail, and $\mathcal{L}_{\mathrm{prior}}$ keeps the prediction close to the pretrained SR model.
Finally, $\mathcal{L}_{\mathrm{ema}}$ adds a slowly varying target only at $s_m$, where supervision ends.


In short, the objectives extend training beyond the supervision boundary and keep using the evidence that remains available. \method{} preserves what can still be verified, guides the remaining fine-scale detail, and constrains that detail with the latent prior.

\subsection{Connection to On-Policy Self-Distillation}
\label{app:opsd_relation}

\method{} is inspired by On-Policy Self-Distillation (OPSD)~\cite{zhao2026selfdistilled}, but differs in how supervision is provided beyond the supervision boundary.
Classical knowledge distillation transfers outputs from a reference model to another model~\cite{hinton2015distilling,rusu2016policy,kim2016sequence}.
Privileged-information methods use additional information that is available during training but not at inference~\cite{lopezpaz2016unifying}.
Other work reduces the mismatch between training and inference.
These methods learn from intermediate predictions or targets that the model itself generates~\cite{ross2011dagger,bengio2015scheduled,lamb2016professor}.
OPSD trains on model-generated sequences while using additional training-only information to supervise those sequences~\cite{zhao2026selfdistilled}.
Related work applies on-policy distillation to vision-language reasoning~\cite{bousselham2026vold}.
Other work applies on-policy self-distillation to multimodal reasoning, fine-detail perception, autoregressive video generation, and diffusion models~\cite{li2026visual,yuan2026vision,liu2026opsdv,zhou2026diffusionopsd}.

\method{} shares two elements with OPSD.
First, it trains on model-generated intermediate predictions.
Second, at the supervision boundary $s_m$, it uses a self-teacher that sees the privileged ground truth.
Unlike OPSD, whose teacher shares the student's parameters, this teacher is an EMA copy.
In recursive SR, each prediction becomes the input to the next scale, but beyond $s_m$ no high-resolution target exists for the deeper prediction.
\method{} uses the last available ground truth as a cross-scale reference for what remains verifiable.
In addition, $\mathcal{L}_{\mathrm{qual}}$ guides the remaining fine-scale detail, and $\mathcal{L}_{\mathrm{prior}}$ limits deviation from the pretrained SR model.
The frozen base model $F_0$ provides the latent target of $\mathcal{L}_{\mathrm{prior}}$.
This target acts as a regularizer, similar in idea to distillation and self-training~\cite{hinton2015distilling,kim2016sequence,xie2020noisy}.
Beyond $s_m$, ground-truth information enters only through $\mathcal{L}_{\mathrm{xscale}}$.
During training, $\mathcal{Q}$ scores the full current prediction $\hat I_j$.
$\mathcal{Q}$ is a no-reference model, so it adds no ground-truth information.
OPSD motivates training on model-generated intermediate predictions.
\method{} adds the cross-scale consistency that recursive SR needs when it continues beyond the supervision boundary.

\section{Implementation Details}
\label{app:implementation}

\begin{table}[t]
\centering
\resizebox{\columnwidth}{!}{%
\begin{tabular}{ll}
\toprule
Hyperparameter & Value \\
\midrule

\multicolumn{2}{l}{\emph{Backbone \& prompter (both frozen)}} \\
SR backbone            & OSEDiff on SD3-medium \\
Prompter (VLM)         & Qwen2.5-VL-3B-Instruct + CoZ GRPO LoRA \\
Prompt length          & at most $32$ tokens \\

\midrule
\multicolumn{2}{l}{\emph{LoRA adapter (only trainable weights)}} \\
Placement              & SD3 transformer (MMDiT) \\
Rank                   & $16$ \\
$\alpha$ (scaling)     & $32$ ($2\times$ rank) \\
Dropout                & $0.0$ \\
Adapted layers & \makecell[l]{query, key, value projections\\(image and text streams)} \\
Trainable parameters   & $7.1$M \\

\midrule
\multicolumn{2}{l}{\emph{Optimization}} \\
Numeric precision      & $32$-bit floating point \\
Optimizer              & AdamW ($\beta_1{=}0.9$, $\beta_2{=}0.999$, $\epsilon{=}10^{-8}$) \\
Learning rate          & $5\times10^{-5}$ \\
Weight decay           & $10^{-2}$ \\
LR schedule            & linear warmup ($500$ steps) then constant \\
Effective batch size   & $4$ (accumulated one image at a time) \\
Max epochs             & $200$ (cap) \\
Early stopping & \makecell[l]{patience $8$ evaluations,\\minimum improvement $10^{-3}$,\\evaluation every $100$ steps} \\
Seed                   & $0$ \\

\midrule
\multicolumn{2}{l}{\emph{Training objectives}} \\
$4\times$ supervision $\lambda_{\mathrm{sup}}$             & $1.0$ \\
Cross-scale consistency $\lambda_{\mathrm{xscale}}$        & $1.0$ \\
Quality objective $\lambda_{\mathrm{qual}}$                & $0.4$ \\
Latent prior $\lambda_{\mathrm{prior}}$     & $8.0$ \\
EMA consistency $\lambda_{\mathrm{ema}}$                   & $0.1$ \\
EMA decay                                                   & $0.95$ \\
Quality model                                               & TOPIQ-NR \\
Latent reference                                            & frozen base model $F_0$ \\

\midrule
\multicolumn{2}{l}{\emph{Data \& protocol}} \\
Training set               & $1{,}000$ curated 4KLSDB images \\
Validation set             & held-out set, $2{,}000$ images \\
Supervision scale          & $4\times$ only (GT) \\
Training recursion         & $4\times \rightarrow 16\times$ \\
Crop size                  & $512\times512$ \\
Recursions at test         & $4$ ($4\times$/$16\times$/$64\times$/$256\times$) \\

\bottomrule
\end{tabular}%
}
\caption{\textbf{Training configuration for \method{}.}
Only the rank-16 LoRA adapter is optimized.
The SR backbone, VAE decoder, VLM prompter, quality model, and $F_0$ remain frozen.}
\label{tab:hyper}
\vspace{-10pt}
\end{table}

\begin{table}[ht]
\centering
\small
\renewcommand{\arraystretch}{0.82}
\setlength{\tabcolsep}{4pt}
\begin{tabular}{lccc}
\toprule
Method & SR (s) & PE (s) & Total (s) \\
\midrule
CoZ (baseline)                & $0.152$ & $1.334$ & $1.486$ \\
\rowcolor{oursbg}
\textbf{\method{}} (separate LoRA) & $0.171$ & $1.334$ & $1.505$ \\
\method{} (merged LoRA)            & $0.153$ & $1.347$ & $1.500$ \\
\bottomrule
\end{tabular}
\caption{
\textbf{Inference latency per recursive step.} Measured on 500 DIV2K images with one H100.
SR: super-resolution pass. PE: prompt extraction by the VLM.
}
\label{tab:latency}
\end{table}
\begin{table}[t]
\centering
\small
\begin{tabular}{@{}lcccc@{}}
\toprule
Test set & $4\times$ & $16\times$ & $64\times$ & $256\times$ \\
\midrule
4KLSDB   & $1016$ & $254$ & $63$ & $16$ \\
DIV8K    & $1104$ & $276$ & $69$ & $17$ \\
DRealSR  & $917$  & $229$ & $57$ & $14$ \\
RealSR   & $287$  & $72$  & $18$ & $4$ \\
Flickr2K & $339$  & $85$  & $21$ & $5$ \\
FFHQ     & $128$  & $32$  & $8$  & $2$ \\
\bottomrule
\end{tabular}
\caption{\textbf{Native side of the judge anchor in pixels.} Median over the judge items of each test set.}
\label{tab:anchor-size}
\vspace{-5pt}
\end{table}

\noindent\textbf{Training Configuration.}
We apply the LoRA adapter to the attention layers of the SD3 transformer.
In each attention layer, the adapter targets the query, key, and value projections of both the image stream and the text stream.
We set the LoRA scaling to $\alpha=32$ and use no dropout.
The prompter, a Vision-Language Model (VLM), generates at most 32 tokens.
AdamW uses $(\beta_1,\beta_2)=(0.9,0.999)$ and $\epsilon=10^{-8}$, with a constant learning rate after warmup.
We evaluate the total training objective on the validation split every 100 optimization steps and use early stopping on it with patience $8$ and a minimum improvement of $10^{-3}$.
Training runs for at most 200 epochs with random seed $0$.
Each micro-batch holds one image, and we accumulate gradients over four micro-batches to reach the effective batch size of $4$.
\Cref{tab:hyper} gives the complete training setup.

\vspace{3pt}
\noindent\textbf{VLM Prompter.}
\label{app:prompter}
The prompter $G$ is Qwen2.5-VL-3B-Instruct with the GRPO-tuned LoRA released by CoZ~\cite{kim2025chain}, and it is frozen in all experiments.
At every recursion step, including the first, the prompter receives one system message and one user turn.
The user turn holds two images in this order: the previous image $\hat I_{i-1}$ (with $\hat I_0{=}I_0$) and its zoomed crop $I_i^{\mathrm{in}}$.
The user turn contains no text. The system message is the CoZ prompt:


\begin{center}
\begin{tcolorbox}[enhanced, width=\linewidth, colback=tzBlueFill, colframe=tzBlueBorder, boxrule=1.2pt, arc=6pt, left=6pt, right=6pt, top=4pt, bottom=4pt,
    title={\small System message of the prompter $G$}, coltitle=white, colbacktitle=tzBlueHeader2, fonttitle=\bfseries]
  \small\setlength{\parindent}{0pt}\raggedright
  The second image is a zoom-in of the first image. Based on this knowledge, what is in the second image? Give me a set of words.
\end{tcolorbox}
\captionof{figure}{\textbf{Prompt of the VLM prompter.} The system message of $G$ at every recursion step, taken from CoZ~\cite{kim2025chain}.}
\label{fig:rubric_prompter}
\end{center}

The decoded answer, at most 32 tokens, is used directly as the SR prompt $p_i$.
During training, the prompter generates the second-step prompt online from the student's own $4\times$ output.
This is the same way the inference pipeline produces prompts.

\vspace{3pt}
\noindent\textbf{Compute and Latency.}
We train on a single node with eight H100-class GPUs.
One GPU runs the optimization, and the other seven GPUs run validation in parallel.
The reported model finishes training in under one day.
To measure latency, we run 500 DIV2K images on one H100 with batch size $1$.
We wait for all GPU work to finish before and after each timed stage, and we discard three warm-up samples.
As shown in \cref{tab:latency}, \method{} requires $1.505$\,s per recursive step compared with $1.486$\,s for CoZ.
The additional cost comes entirely from the LoRA-adapted SR pass.
Prompt extraction (PE in \cref{tab:latency}) remains unchanged.
Since each recursive input is resampled to $512{\times}512$, the per-step cost remains approximately constant from $4{\times}$ to $256{\times}$.
When we merge the linear LoRA weights into the backbone, the SR latency is within $0.001$\,s of CoZ.

\vspace{3pt}
\noindent\textbf{Inputs and Metrics.}
The first model input is the center quarter ($128$ pixels) of $I_0$, resized to $512{\times}512$ with bicubic interpolation.
The $4\times$ training target is the native center square of side $S/4$, where $S$ is the short side of the image, resized to $512{\times}512$ with LANCZOS.
FFHQ ($512{\times}512$ version) and Flickr2K are evenly spaced subsets.
4KLSDB uses the first $200$ test images, in stored order, with a short side of at least $3840$ pixels, a photographic caption, and no near-duplicate in DIV2K.
DIV8K uses all $1{,}238$ images of the official training split with a short side of at least $2048$ pixels. \method{} does not train on DIV8K.
RealSR uses the $100$ test images of Version 3 ($50$ Canon and $50$ Nikon).
We compute the no-reference metrics with \texttt{pyiqa} $0.1.15$. MUSIQ and TOPIQ-NR use KonIQ-10k checkpoints, MANIQA uses the PIPAL checkpoint, and CLIPIQA uses RN50 CLIP without learned prompts.

\vspace{3pt}
\noindent\textbf{Judge Protocol.}
Both judges (Gemini 4 Argon and GPT-6.1 Sol) receive the same prompt and images.
Each item contains the native ground-truth window that the zoom step magnifies (the anchor) and two candidates of the same region.
Every pair is judged in both A/B orders.
The judges are \texttt{gemini-4-argon} and \texttt{gpt-6.1-sol}, queried in November 2026. Gemini 4 Argon uses temperature $0$. GPT-6.1 Sol uses its default temperature, because its API does not accept this setting.
The anchor is the center square of side $\mathrm{round}(S/4^{i})$ native pixels, enlarged to $512{\times}512$ with LANCZOS (\cref{tab:anchor-size}).
Each image carries an explicit \texttt{ANCHOR}, \texttt{A}, or \texttt{B} label.
The output contains a comparison in $\{\texttt{A},\texttt{B},\texttt{tie}\}$ and one true/false hallucination flag for each candidate.
We record unparsable outputs as abstentions.
Win rates use only decided comparisons, that is, judgments with no tie and no abstention.
\Cref{tab:gemini-judge} reports the rates and counts.
The exact prompt is in \cref{app:prompts}.

\FloatBarrier
\section{Additional Results and Analysis}
\label{app:additional_results}

\subsection{Additional Ablations and Analysis}
\label{app:additional_analysis}

\begin{table*}[t]
    \centering
    \small
    \setlength{\tabcolsep}{3.2pt}
    \begin{tabular}{l cccc cccc cccc}
        \toprule
        & \multicolumn{4}{c}{CLIPIQA$\uparrow$} & \multicolumn{4}{c}{MUSIQ$\uparrow$} & \multicolumn{4}{c}{Hallucination rate$\downarrow$} \\
        \cmidrule(lr){2-5}\cmidrule(lr){6-9}\cmidrule(l){10-13}
        Variant & $4\times$ & $16\times$ & $64\times$ & $256\times$ & $4\times$ & $16\times$ & $64\times$ & $256\times$ & $4\times$ & $16\times$ & $64\times$ & $256\times$ \\
        \midrule
        \rowcolor{oursbg} Full \method{} & 0.711 & 0.710 & 0.714 & 0.707 & 65.1 & 58.5 & 54.6 & 52.2 & 0.37 & 0.45 & 0.53 & 0.68 \\
        $-\mathcal{L}_{\mathrm{sup}}$ & 0.744 & 0.719 & 0.705 & 0.691 & 68.8 & 60.5 & 56.6 & 54.4 & 0.57 & 0.63 & 0.68 & 0.68 \\
        $-\mathcal{L}_{\mathrm{xscale}}$ & 0.706 & 0.727 & 0.723 & 0.714 & 65.7 & 60.9 & 56.6 & 53.8 & 0.39 & 0.59 & 0.69 & 0.80 \\
        $-\mathcal{L}_{\mathrm{qual}}$ & 0.659 & 0.594 & 0.591 & 0.577 & 62.3 & 52.7 & 49.0 & 47.7 & 0.35 & 0.34 & 0.41 & 0.52 \\
        $-\mathcal{L}_{\mathrm{prior}}$ & 0.610 & 0.794 & 0.799 & 0.822 & 58.8 & 59.8 & 64.1 & 64.4 & 0.50 & 1.00 & 0.99 & 0.99 \\
        $-\mathcal{L}_{\mathrm{ema}}$ & 0.716 & 0.702 & 0.707 & 0.698 & 65.7 & 58.8 & 55.0 & 52.4 & 0.36 & 0.44 & 0.53 & 0.70 \\
        \midrule
        Fine-tuning only & 0.621 & 0.453 & 0.421 & 0.440 & 58.8 & 40.1 & 36.5 & 40.6 & 0.27 & 0.27 & 0.25 & 0.30 \\
        Detached recursion & 0.700 & 0.709 & 0.718 & 0.714 & 64.9 & 58.2 & 54.4 & 52.2 & 0.35 & 0.44 & 0.55 & 0.64 \\
        \bottomrule
    \end{tabular}%
    \caption{\textbf{Ablation of the training objectives at every scale.}
    The variants of Table 3 in the main paper, scored at all four zoom depths.
    We average metrics over the six test sets other than DIV2K.
    Hallucination rate: ground-truth-anchored judges, pooled over Gemini 4 Argon and GPT-6.1 Sol ($480$ judgments per scale).}
    \label{tab:abl-depth}
\end{table*}

\begin{table*}[t]
    \centering
    \small
    \setlength{\tabcolsep}{4pt}
    \begin{tabular}{l cc ccc c c c}
        \toprule
        & \multicolumn{2}{c}{CLIPIQA$\uparrow$}
        & \multicolumn{3}{c}{Fidelity @$4\times$ (GT)}
        & & & \\
        \cmidrule(lr){2-3}
        \cmidrule(lr){4-6}
        Quality model
        & Mean
        & $256\times$
        & LPIPS$\downarrow$
        & DISTS$\downarrow$
        & DINOv2$\uparrow$
        & Coherence$\uparrow$
        & TOPIQ$_{256}\uparrow$
        & Halluc.$\downarrow$ \\
        \midrule
        \rowcolor{oursbg}
        \textbf{\method{}} (TOPIQ-NR)
        & \textbf{0.711}
        & \textbf{0.707}
        & 0.214
        & 0.160
        & 0.926
        & 0.792
        & \textbf{0.585}
        & 0.51 \\
        \midrule
        HyperIQA
        & 0.680\,\loss{$-0.031$}
        & 0.680\,\loss{$-0.027$}
        & 0.214
        & 0.159
        & \textbf{0.928}
        & \textbf{0.816}\,\gain{$+0.024$}
        & 0.553
        & 0.61 \\
        Laplacian
        & 0.610\,\loss{$-0.100$}
        & 0.579\,\loss{$-0.128$}
        & 0.214
        & \textbf{0.158}
        & 0.926
        & 0.808\,\gain{$+0.016$}
        & 0.467
        & \textbf{0.40} \\
        \bottomrule
    \end{tabular}
    \caption{
\textbf{Choice of no-reference quality model.}
    We average no-reference metrics over the six test sets other than DIV2K. GT fidelity counts all $1{,}631$ images of the four sets with $4\times$ targets equally.
    Coherence: cosine similarity of DINOv2 CLS features between consecutive zooms. Halluc.: rate of the ground-truth-anchored judges, pooled over both judges and the four scales.
}
    \label{tab:abl-reward-multi}
\end{table*}

\begin{table*}[t]
    \centering
    \small
    \resizebox{\textwidth}{!}{%
        \begin{tabular}{l ccc ccc ccc ccc}
            \toprule
            & \multicolumn{3}{c}{\textbf{4KLSDB}} & \multicolumn{3}{c}{\textbf{DIV8K}} & \multicolumn{3}{c}{\textbf{DRealSR}} & \multicolumn{3}{c}{\textbf{RealSR}} \\
            \cmidrule(lr){2-4} \cmidrule(lr){5-7} \cmidrule(lr){8-10} \cmidrule(lr){11-13}
            Method & LPIPS$\downarrow$ & DISTS$\downarrow$ & DINOv2$\uparrow$ & LPIPS$\downarrow$ & DISTS$\downarrow$ & DINOv2$\uparrow$ & LPIPS$\downarrow$ & DISTS$\downarrow$ & DINOv2$\uparrow$ & LPIPS$\downarrow$ & DISTS$\downarrow$ & DINOv2$\uparrow$ \\
            \midrule
            HiT-SR~\cite{zhang2024hierarchical} & 0.408 & 0.232 & 0.907 & 0.448 & 0.265 & 0.894 & 0.322 & 0.223 & 0.920 & 0.186 & 0.169 & \underline{0.927} \\
            MambaIR~\cite{guo2024mambair} & 0.412 & 0.233 & 0.903 & 0.449 & 0.268 & 0.897 & 0.326 & 0.225 & 0.916 & 0.185 & 0.169 & \textbf{0.928} \\
            SwinIR~\cite{liang2021swinir} & 0.280 & 0.184 & \textbf{0.927} & 0.320 & 0.210 & \underline{0.919} & 0.197 & 0.174 & \textbf{0.930} & \textbf{0.115} & \textbf{0.131} & 0.886 \\
            SeeSR~\cite{wu2024seesr} & 0.260 & \underline{0.174} & 0.917 & 0.263 & 0.174 & 0.917 & \textbf{0.177} & \textbf{0.156} & \underline{0.928} & \underline{0.164} & \underline{0.151} & 0.885 \\
            OSEDiff~\cite{wu2024one} & 0.335 & 0.215 & 0.793 & 0.380 & 0.243 & 0.788 & 0.318 & 0.231 & 0.805 & 0.310 & 0.223 & 0.729 \\
            CoZ~\cite{kim2025chain} & \underline{0.248} & 0.186 & 0.887 & \underline{0.238} & \underline{0.167} & 0.910 & 0.183 & \underline{0.158} & 0.900 & 0.193 & 0.170 & 0.860 \\
            \rowcolor{oursbg} \textbf{\method{}} & \textbf{0.169} & \textbf{0.142} & \underline{0.925} & \textbf{0.223} & \textbf{0.161} & \textbf{0.930} & \underline{0.182} & 0.161 & 0.917 & 0.220 & 0.176 & 0.890 \\
            \bottomrule
        \end{tabular}%
    }
    \caption{\textbf{Full-reference fidelity at $4\times$.} Results on the four datasets with high-resolution targets.
    \method{} performs best on 4KLSDB and DIV8K, while results are more mixed on the real-camera DRealSR and RealSR sets.
    \textbf{Bold}: best. \underline{Underline}: second-best.}
    \label{tab:fidelity-perdataset}
\end{table*}
\begin{table}[t]
    \centering
    \footnotesize
    \setlength{\tabcolsep}{1.6pt}
        \begin{tabular}{lccccccc}
            \toprule
            & \multicolumn{4}{c}{CLIPIQA$\uparrow$} & MUSIQ$\uparrow$ & \multicolumn{2}{c}{Fidelity @$4\times$} \\
            \cmidrule(lr){2-5}\cmidrule(lr){7-8}
            Prompt & $4\times$ & $16\times$ & $64\times$ & $256\times$ & $256\times$ & LPIPS$\downarrow$ & DISTS$\downarrow$ \\
            \midrule
            \rowcolor{oursbg} GRPO VLM & $\mathbf{0.714}$ & $0.714$ & $\mathbf{0.716}$ & $\mathbf{0.706}$ & $\mathbf{52.1}$ & $\mathbf{0.214}$ & $\mathbf{0.160}$ \\
            Base VLM & $0.712$ & $0.715$ & $0.712$ & $0.697$ & $50.5$ & $\mathbf{0.214}$ & $\mathbf{0.160}$ \\
            No prompt & $0.710$ & $\mathbf{0.727}$ & $0.708$ & $0.673$ & $47.8$ & $0.219$ & $0.166$ \\
            \bottomrule
        \end{tabular}
    \caption{
\textbf{Effect of the text prompt at inference.}
    The same trained adapter with three prompt sources: the GRPO-tuned Qwen2.5-VL-3B prompter of CoZ~\cite{kim2025chain} (default), the base Qwen2.5-VL-3B (no GRPO LoRA), and an empty prompt.
    We average no-reference metrics over the seven test sets. GT fidelity counts all $1{,}631$ images of the four sets with $4\times$ targets equally.
}
    \label{tab:prompt-ablation}
\end{table}

\begin{table}[t]
    \centering
    \scriptsize
    \setlength{\tabcolsep}{2.9pt}
    \begin{tabular}{@{}lcccccc@{}}
        \toprule
        & \multicolumn{3}{c}{\textbf{Gemini 4 Argon}} & \multicolumn{3}{c}{\textbf{GPT-6.1 Sol}} \\
        \cmidrule(lr){2-4}\cmidrule(l){5-7}
        \textbf{Scale} & \method{} & CoZ & Pref. & \method{} & CoZ & Pref. \\
        \midrule
        $4\times$   & $\mathbf{0.40}$ & $0.45$ & $86$/$119$/$35$ & $\mathbf{0.34}$ & $0.39$ & $67$/$140$/$33$ \\
        $16\times$  & $\mathbf{0.44}$ & $0.46$ & $88$/$71$/$81$ & $0.47$ & $\mathbf{0.44}$ & $89$/$71$/$80$ \\
        $64\times$  & $\mathbf{0.49}$ & $0.53$ & $96$/$60$/$84$ & $0.56$ & $\mathbf{0.53}$ & $93$/$63$/$84$ \\
        $256\times$ & $0.65$ & $\mathbf{0.57}$ & $80$/$62$/$98$ & $0.71$ & $\mathbf{0.63}$ & $95$/$68$/$77$ \\
        \bottomrule
    \end{tabular}
    \caption{
\textbf{Ground-truth-anchored judges: \method{} vs.\ CoZ.}
    Hallucination rate of each method, and preference counts (\method{} / tie / CoZ).
    Each scale has $120$ items ($20$ per test set except DIV2K), each judged in both A/B orders ($240$ judgments per judge).
}
    \label{tab:gemini-judge}
\end{table}

\begin{table}[t]
    \centering
    \small
    \setlength{\tabcolsep}{4pt}
    \resizebox{\columnwidth}{!}{%
    \begin{tabular}{@{}ccccc@{}}
        \toprule
        Scale & Reference & \method{} & CoZ & \method{} closer \\
        \midrule
        $4\times$  & $512$ px & $\mathbf{0.00588}$ & $0.00599$ & $80$ / $120$ \\
        $16\times$ & $128$ px & $\mathbf{0.00891}$ & $0.00985$ & $68$ / $120$ \\
        \bottomrule
    \end{tabular}%
    }
    \caption{
\textbf{Gromov-Wasserstein distance to the last ground truth (lower is better).}
    Each output is compared with the region it magnifies, using the structure of DINOv2 patch features.
}
    \label{tab:gw}
\end{table}

\begin{table}[t]
\centering
\small
\setlength{\tabcolsep}{5.8pt}
\begin{tabular}{@{}cccc@{}}
\toprule
Scale & Flag agreement & Cohen's $\kappa$ & Same flag in both orders \\
\midrule
$4\times$   & $0.83$ & $0.64$ & $0.86$ / $0.90$ \\
$16\times$  & $0.77$ & $0.53$ & $0.89$ / $0.80$ \\
$64\times$  & $0.74$ & $0.48$ & $0.82$ / $0.83$ \\
$256\times$ & $0.78$ & $0.51$ & $0.84$ / $0.91$ \\
\bottomrule
\end{tabular}
\caption{\textbf{Agreement of the ground-truth-anchored judges.}
Flag agreement and Cohen's $\kappa$ compare the hallucination flags of the two judges. The last column: Gemini 4 Argon / GPT-6.1 Sol.}
\label{tab:judge-stats}
\end{table}

We further examine which no-reference quality model to use.
We also report the ground-truth-anchored judges, the reliability of the quality model, a Gromov-Wasserstein comparison, the text-prompt effect, and per-dataset fidelity.

\vspace{3pt}
\noindent\textbf{Ablation at Every Scale.}
\Cref{tab:abl-depth} scores the ablation variants of the main paper at all four scales.
The full model keeps CLIPIQA near $0.71$ from $4\times$ to $256\times$.
Removing $\mathcal{L}_{\mathrm{sup}}$ or $\mathcal{L}_{\mathrm{xscale}}$ raises MUSIQ at every scale.
The hallucination rate rises as well, with one exception: without $\mathcal{L}_{\mathrm{sup}}$ it stays at $0.68$ at $256\times$.
Without $\mathcal{L}_{\mathrm{xscale}}$ it reaches $0.80$ at $256\times$, against $0.68$ for the full model.
Two variants, one without $\mathcal{L}_{\mathrm{qual}}$ and one with fine-tuning only, stay low in quality at every depth. Their outputs contradict the scene less often, because they add less detail.
Without $\mathcal{L}_{\mathrm{prior}}$, CLIPIQA keeps rising ($0.822$ at $256\times$) while the hallucination rate stays near $1.00$ from $16\times$ on, which is reward hacking.
Removing $\mathcal{L}_{\mathrm{ema}}$ changes CLIPIQA, MUSIQ, and the hallucination rate only slightly.
This agrees with its role in the main paper as a minor term at the supervision boundary.
Detached recursion stays near the full model at all scales.

\begin{figure*}[t]
\centering
\includegraphics[width=0.9\linewidth]{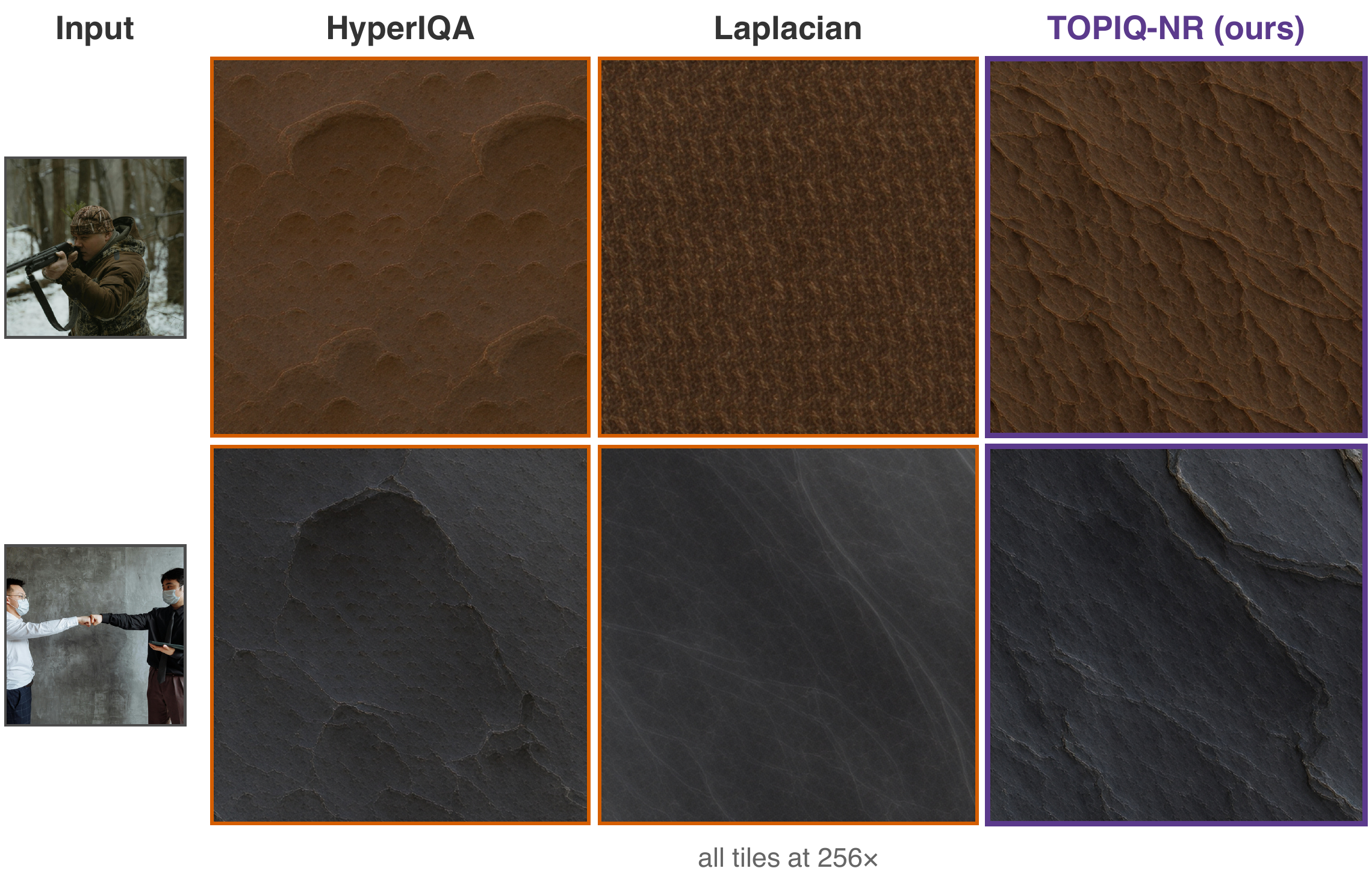}
\caption{
\textbf{Effect of the no-reference quality model at $256\times$.}
Two 4KLSDB examples with all other settings fixed.
HyperIQA repeats rounded blob shapes, and Laplacian sharpness gives knit-like noise or faint streaks.
TOPIQ-NR, used by \method{}, gives irregular, natural detail (see \cref{tab:abl-reward-multi}).
}
\label{fig:reward-critic}
\end{figure*}

\vspace{3pt}
\noindent\textbf{Choice of Quality Model.}
\Cref{tab:abl-reward-multi} replaces TOPIQ-NR, the quality model that supplies $\mathcal{L}_{\mathrm{qual}}$ in training, with HyperIQA~\cite{su2020hyperiqa} or a Laplacian sharpness measure.
TOPIQ-NR gives the highest CLIPIQA, both as a mean and at $256{\times}$.
All three choices give similar $4{\times}$ fidelity.
HyperIQA slightly improves coherence.
However, it reduces mean CLIPIQA by $0.031$ and CLIPIQA at $256{\times}$ by $0.027$.
The Laplacian variant reduces mean CLIPIQA by $0.100$ and CLIPIQA at $256{\times}$ by $0.128$.
HyperIQA hallucinates more often than TOPIQ-NR ($0.61$ vs.\ $0.51$), and the Laplacian variant hallucinates less often ($0.40$) because it adds less detail.
\Cref{fig:reward-critic} shows qualitative examples for the same three quality models.
In these examples, the Laplacian variant loses fine structure, and HyperIQA produces less realistic textures.
HyperIQA has slightly better coherence in \cref{tab:abl-reward-multi}.
However, at $256\times$ it repeats the same blob shapes in \cref{fig:reward-critic}, and \method{} does not.

\vspace{3pt}
\noindent\textbf{Ground-Truth-Anchored Judges.}
\Cref{tab:gemini-judge} gives the hallucination rates and preference counts of both judges.
\Cref{tab:judge-stats} gives the agreement between the two judges and the consistency of each judge across A/B orders.
Both prefer \method{} at $4\times$ and flag it less often than CoZ.
At $16\times$ and $64\times$, the two methods are comparable, and the judges differ on which one is flagged less.
At $256\times$, both judges flag \method{} more often. However, the anchor holds very few native pixels.
When both judges pick a winner, they pick the same one in $366$ of $490$ cases ($75\%$).
Neither judge favors position A or B ($52\%$ and $47\%$ of decided comparisons pick A).
In a blind study, two raters compared \method{} and CoZ on $40$ image pairs against the ground-truth window.
\method{} received $80\%$ of the non-tie votes ($48$ vs.\ $12$, with $20$ ties).
The raters were volunteer PhD students in computer science, not authors of this paper, who gave informed consent to take part and to share their ratings.
We did not seek institutional review, because the study collected only image preferences and no personal data.

\FloatBarrier

\begin{figure}[t]
    \centering
    \includegraphics[width=\linewidth]{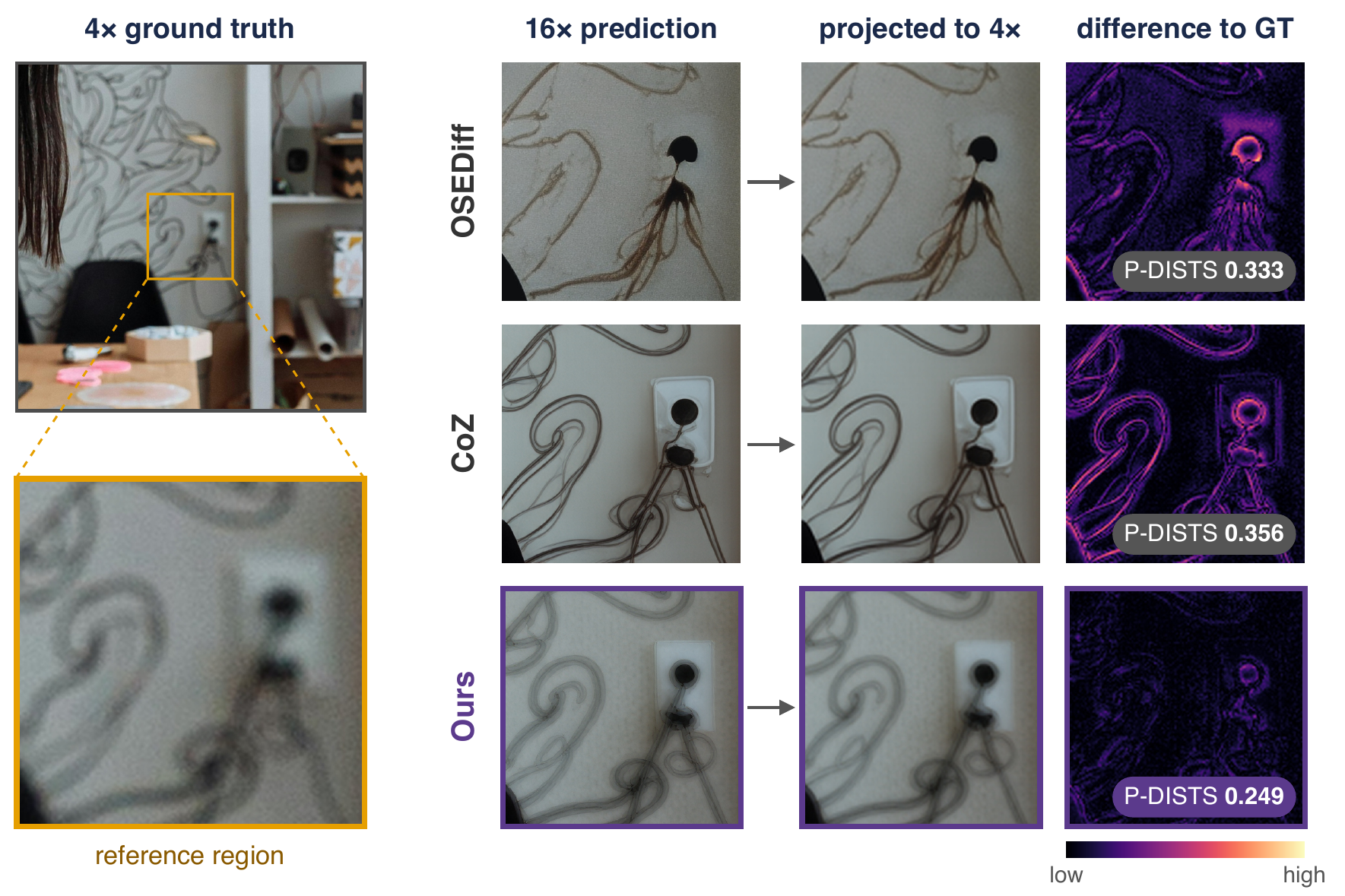}
    \caption{
\textbf{Projected fidelity at $16\times$.}
    Left: the $4\times$ ground truth, with the region that the $16\times$ zoom magnifies (boxed, enlarged below).
    Each $16\times$ prediction is downsampled to $4\times$ and compared with this region.
    CoZ redraws the soft cords as hard double lines, and OSEDiff turns the socket into a black blob with root-like strands. \method{} stays closest to the reference, which is soft in this region.
}
    \label{fig:projection_reveal}
\end{figure}

\vspace{3pt}
\noindent\textbf{Judge Validation.}
We test both judges on known answers, using $60$ judge items from 4KLSDB, DIV8K, and DRealSR ($20$ each) in both A/B orders.
In the first test, the anchor and one candidate are a clean image, and the other candidate is that image with one planted artifact (eight types, patch side $32$, $64$, or $128$ pixels).
The judges flag the artifact in $77\%$ to $96\%$ of judgments, and they flag the clean image in $0$ of $5{,}760$ judgments.
In the second test, one candidate is the real $4\times$ target and the other is a $4\times$ output of \method{} or CoZ.
The anchor is this target, at native size or reduced.
With the native anchor or an anchor reduced to $256$ pixels, the judges flag the real image in at most $0.4\%$ of judgments and the model output in $86\%$ to $92\%$.
With an anchor reduced to $64$ pixels, they no longer separate the real image from the model output.
We therefore treat results with small anchors as weak evidence (\cref{tab:anchor-size}).
Flag rates depend on the paired candidate, so we compare methods only against the same opponent.

\vspace{3pt}
\noindent\textbf{Reliability of the Quality Model.}
We add eight artifact types to $120$ clean $512{\times}512$ images: checkerboard, stripes, speckle, mosaic blocks, broken edges, repeated texture, blur, and pasted regions. Each artifact fills one patch of side $32$, $64$, or $128$ pixels.
The images are real ground-truth windows and $16\times$ outputs of \method{}.
The TOPIQ-NR score drops in only $46\%$ to $60\%$ of cases, depending on patch size and image source. Its mean change is at most $0.14$ standard deviations.
At side $128$, stripes raise the score in $62.5\%$ of cases, and mosaic blocks on $16\times$ outputs lower it in $83\%$.
Pasted regions, repeated texture, and broken edges at side $128$ remain detectable after the $4\times$ bicubic downsampling used in $\mathcal{L}_{\mathrm{xscale}}$. Their LPIPS to the clean image after downsampling is at least $79\%$ of its value at full resolution.
At the same side, stripes and speckle mostly vanish after downsampling, and this ratio is at most $13\%$.
The score alone is not a reliable check on small artifacts.
$\mathcal{L}_{\mathrm{xscale}}$ can check only artifacts that survive projection.
Training thus uses $\mathcal{L}_{\mathrm{prior}}$ to limit drift from $F_0$.

\vspace{3pt}
\noindent\textbf{Projected Fidelity Example.}
\Cref{fig:projection_reveal} shows one image from the $16\times$ projected-fidelity evaluation.
The main paper reports the mean P-DISTS over all $1{,}631$ images of the four sets with $4\times$ targets.

\vspace{3pt}
\noindent\textbf{Gromov-Wasserstein Comparison.}
This distance compares the internal structure of two feature sets, so it needs no pixel alignment.
We compute it on DINOv2 patch features between each output and the ground-truth region it magnifies, on the $120$ judge items (\cref{tab:gw}).
\method{} is closer to the ground truth at $4\times$ and at $16\times$.
As a sanity check, an output is closer to its own reference than to that of another image in $100\%$ of cases at $4\times$ and $88\%$ at $16\times$.

\vspace{3pt}
\noindent\textbf{Text Prompt.}
\label{app:prompt_ablation}
\Cref{tab:prompt-ablation} varies only the prompt source at inference.
The prompt matters most at the deepest zoom.
With an empty prompt, CLIPIQA at $256{\times}$ drops from $0.706$ to $0.673$, and MUSIQ at $256{\times}$ drops from $52.1$ to $47.8$.
In contrast, $4{\times}$ fidelity stays almost the same.
The base Qwen2.5-VL-3B without the GRPO LoRA performs close to the tuned prompter ($0.697$ vs.\ $0.706$ CLIPIQA at $256{\times}$).
Even with an empty prompt, \method{} stays well above CoZ with its full prompter at $256{\times}$ ($0.673$ vs.\ $0.579$ CLIPIQA).
Thus, the adapter gives most of the gain.

\vspace{3pt}
\noindent\textbf{Per-Dataset Fidelity.}
\Cref{tab:fidelity-perdataset} expands the $4{\times}$ fidelity evaluation to all four datasets with high-resolution targets.
\method{} gives the strongest LPIPS and DISTS on 4KLSDB and the strongest LPIPS, DISTS, and DINOv2 similarity on DIV8K.
Results are more mixed on DRealSR and RealSR.
The largest fidelity gains occur on the high-resolution photographic sets, 4KLSDB and DIV8K.
The aggregate counts every image equally, so DIV8K contributes $76\%$ of it. Without the in-domain 4KLSDB set, mean LPIPS is $0.220$ for \method{}, $0.231$ for CoZ, and $0.250$ for SeeSR, so the lead holds on out-of-domain data, mainly DIV8K.
A native $4\times$ target needs a short side of at least $2048$ pixels. All images of 4KLSDB, DIV8K, and DRealSR meet this, but only $4$ of the $100$ RealSR images do, so the other $96$ RealSR targets are upsampled.
Without RealSR, \method{} still ranks first: LPIPS is $0.214$ against $0.236$ for CoZ, DISTS is $0.159$ against $0.168$ for CoZ, and P-DISTS at $16\times$ is $0.212$ against $0.223$ for SeeSR.

\FloatBarrier
\subsection{Per-Dataset Results}
\label{app:results_breakdown}
This subsection extends the aggregate results of the main paper.
It gives quantitative results for each dataset and qualitative recursive zoom examples on all seven test sets.

\begin{table}[t]
    \centering
    \small
    \renewcommand{\arraystretch}{0.9}
    \setlength{\tabcolsep}{4pt}
    \resizebox{\columnwidth}{!}{%
        \begin{tabular}{ll cccc}
            \toprule
            Scale & Method & NIQE$\downarrow$ & MUSIQ$\uparrow$ & MANIQA$\uparrow$ & CLIPIQA$\uparrow$ \\
            \midrule
            \multirow{7}{*}{$4\times$}
                & HiT-SR~\cite{zhang2024hierarchical} & 8.76 & 40.1 & 0.396 & 0.396 \\
                & MambaIR~\cite{guo2024mambair} & 8.68 & 41.2 & 0.405 & 0.412 \\
                & SwinIR~\cite{liang2021swinir} & 6.02 & 51.3 & 0.500 & 0.465 \\
                & SeeSR~\cite{wu2024seesr} & \underline{5.44} & \underline{63.5} & \underline{0.569} & 0.615 \\
                & OSEDiff~\cite{wu2024one} & \textbf{5.00} & 60.3 & 0.559 & 0.608 \\
                & CoZ~\cite{kim2025chain} & 5.74 & \textbf{64.4} & \textbf{0.582} & \underline{0.623} \\
            \rowcolor{oursbg}
                & \textbf{\method{}} & 6.00 & 59.8 & 0.556 & \textbf{0.638} \\
            \midrule
            \multirow{7}{*}{$16\times$}
                & HiT-SR~\cite{zhang2024hierarchical} & 14.13 & 19.7 & 0.333 & 0.370 \\
                & MambaIR~\cite{guo2024mambair} & 14.42 & 20.4 & 0.333 & 0.390 \\
                & SwinIR~\cite{liang2021swinir} & 7.34 & 28.4 & 0.427 & 0.454 \\
                & SeeSR~\cite{wu2024seesr} & 7.81 & 47.8 & 0.507 & 0.552 \\
                & OSEDiff~\cite{wu2024one} & \textbf{6.10} & 51.9 & 0.531 & \underline{0.586} \\
                & CoZ~\cite{kim2025chain} & 8.08 & \underline{52.3} & \underline{0.551} & 0.578 \\
            \rowcolor{oursbg}
                & \textbf{\method{}} & \underline{7.16} & \textbf{54.8} & \textbf{0.556} & \textbf{0.658} \\
            \midrule
            \multirow{7}{*}{$64\times$}
                & HiT-SR~\cite{zhang2024hierarchical} & 16.69 & 23.8 & 0.388 & 0.455 \\
                & MambaIR~\cite{guo2024mambair} & 17.82 & 23.3 & 0.388 & 0.469 \\
                & SwinIR~\cite{liang2021swinir} & 9.02 & 26.5 & 0.489 & 0.477 \\
                & SeeSR~\cite{wu2024seesr} & 9.83 & 37.3 & 0.495 & 0.497 \\
                & OSEDiff~\cite{wu2024one} & \textbf{7.20} & \underline{46.2} & 0.531 & 0.533 \\
                & CoZ~\cite{kim2025chain} & 9.54 & 45.7 & \underline{0.555} & \underline{0.544} \\
            \rowcolor{oursbg}
                & \textbf{\method{}} & \underline{8.47} & \textbf{50.8} & \textbf{0.579} & \textbf{0.651} \\
            \midrule
            \multirow{7}{*}{$256\times$}
                & HiT-SR~\cite{zhang2024hierarchical} & 17.81 & 26.2 & 0.428 & 0.494 \\
                & MambaIR~\cite{guo2024mambair} & 18.97 & 26.0 & 0.426 & 0.509 \\
                & SwinIR~\cite{liang2021swinir} & 10.77 & 28.5 & 0.495 & 0.467 \\
                & SeeSR~\cite{wu2024seesr} & 11.56 & 33.7 & 0.503 & 0.496 \\
                & OSEDiff~\cite{wu2024one} & \textbf{8.18} & 42.6 & 0.525 & 0.502 \\
                & CoZ~\cite{kim2025chain} & 10.41 & \underline{44.1} & \underline{0.555} & \underline{0.538} \\
            \rowcolor{oursbg}
                & \textbf{\method{}} & \underline{9.31} & \textbf{49.3} & \textbf{0.589} & \textbf{0.664} \\
            \bottomrule
        \end{tabular}%
    }
    \caption{\textbf{Per-scale no-reference quality on 4KLSDB.}
    \textbf{Bold}: best. \underline{Underline}: second-best.}
    \label{tab:ds_4klsdb}
\end{table}

\begin{table}[t]
    \centering
    \small
    \renewcommand{\arraystretch}{0.9}
    \setlength{\tabcolsep}{4pt}
    \resizebox{\columnwidth}{!}{%
        \begin{tabular}{ll cccc}
            \toprule
            Scale & Method & NIQE$\downarrow$ & MUSIQ$\uparrow$ & MANIQA$\uparrow$ & CLIPIQA$\uparrow$ \\
            \midrule
            \multirow{7}{*}{$4\times$}
                & HiT-SR~\cite{zhang2024hierarchical} & 7.89 & 39.2 & 0.403 & 0.374 \\
                & MambaIR~\cite{guo2024mambair} & 7.81 & 40.6 & 0.419 & 0.386 \\
                & SwinIR~\cite{liang2021swinir} & 5.66 & 51.0 & 0.510 & 0.458 \\
                & SeeSR~\cite{wu2024seesr} & 4.70 & 64.3 & 0.608 & 0.619 \\
                & OSEDiff~\cite{wu2024one} & 4.93 & 59.5 & 0.585 & 0.638 \\
                & CoZ~\cite{kim2025chain} & \underline{4.65} & \textbf{66.9} & \underline{0.633} & \underline{0.699} \\
            \rowcolor{oursbg}
                & \textbf{\method{}} & \textbf{4.46} & \underline{66.5} & \textbf{0.638} & \textbf{0.735} \\
            \midrule
            \multirow{7}{*}{$16\times$}
                & HiT-SR~\cite{zhang2024hierarchical} & 13.11 & 18.4 & 0.310 & 0.312 \\
                & MambaIR~\cite{guo2024mambair} & 13.28 & 19.2 & 0.312 & 0.329 \\
                & SwinIR~\cite{liang2021swinir} & 6.78 & 29.0 & 0.391 & 0.401 \\
                & SeeSR~\cite{wu2024seesr} & 6.50 & 51.8 & 0.523 & 0.540 \\
                & OSEDiff~\cite{wu2024one} & \underline{5.96} & 53.0 & 0.545 & 0.605 \\
                & CoZ~\cite{kim2025chain} & 6.29 & \underline{58.5} & \underline{0.595} & \underline{0.652} \\
            \rowcolor{oursbg}
                & \textbf{\method{}} & \textbf{5.82} & \textbf{61.6} & \textbf{0.599} & \textbf{0.737} \\
            \midrule
            \multirow{7}{*}{$64\times$}
                & HiT-SR~\cite{zhang2024hierarchical} & 16.52 & 21.8 & 0.367 & 0.408 \\
                & MambaIR~\cite{guo2024mambair} & 17.10 & 21.6 & 0.366 & 0.419 \\
                & SwinIR~\cite{liang2021swinir} & 8.09 & 23.2 & 0.478 & 0.465 \\
                & SeeSR~\cite{wu2024seesr} & 8.98 & 42.0 & 0.502 & 0.517 \\
                & OSEDiff~\cite{wu2024one} & \underline{7.30} & 47.7 & 0.536 & 0.576 \\
                & CoZ~\cite{kim2025chain} & 7.86 & \underline{52.1} & \underline{0.581} & \underline{0.623} \\
            \rowcolor{oursbg}
                & \textbf{\method{}} & \textbf{7.12} & \textbf{56.0} & \textbf{0.597} & \textbf{0.727} \\
            \midrule
            \multirow{7}{*}{$256\times$}
                & HiT-SR~\cite{zhang2024hierarchical} & 17.64 & 26.3 & 0.421 & 0.480 \\
                & MambaIR~\cite{guo2024mambair} & 18.47 & 26.2 & 0.418 & 0.493 \\
                & SwinIR~\cite{liang2021swinir} & 9.96 & 27.4 & 0.504 & 0.466 \\
                & SeeSR~\cite{wu2024seesr} & 11.16 & 36.3 & 0.502 & 0.497 \\
                & OSEDiff~\cite{wu2024one} & \underline{8.53} & 43.7 & 0.528 & 0.543 \\
                & CoZ~\cite{kim2025chain} & 9.24 & \underline{48.3} & \underline{0.576} & \underline{0.599} \\
            \rowcolor{oursbg}
                & \textbf{\method{}} & \textbf{8.30} & \textbf{51.8} & \textbf{0.595} & \textbf{0.703} \\
            \bottomrule
        \end{tabular}%
    }
    \caption{\textbf{Per-scale no-reference quality on DIV2K.}
    \textbf{Bold}: best. \underline{Underline}: second-best.}
    \label{tab:ds_div2k}
\end{table}

\begin{table}[t]
    \centering
    \small
    \renewcommand{\arraystretch}{0.9}
    \setlength{\tabcolsep}{4pt}
    \resizebox{\columnwidth}{!}{%
        \begin{tabular}{ll cccc}
            \toprule
            Scale & Method & NIQE$\downarrow$ & MUSIQ$\uparrow$ & MANIQA$\uparrow$ & CLIPIQA$\uparrow$ \\
            \midrule
            \multirow{7}{*}{$4\times$}
                & HiT-SR~\cite{zhang2024hierarchical} & 8.05 & 37.1 & 0.395 & 0.366 \\
                & MambaIR~\cite{guo2024mambair} & 8.00 & 38.5 & 0.412 & 0.377 \\
                & SwinIR~\cite{liang2021swinir} & 5.85 & 49.4 & 0.512 & 0.447 \\
                & SeeSR~\cite{wu2024seesr} & \underline{4.66} & 63.9 & 0.612 & 0.610 \\
                & OSEDiff~\cite{wu2024one} & 4.88 & 60.6 & 0.596 & 0.639 \\
                & CoZ~\cite{kim2025chain} & 4.67 & \underline{67.0} & \underline{0.638} & \underline{0.697} \\
            \rowcolor{oursbg}
                & \textbf{\method{}} & \textbf{4.48} & \textbf{67.3} & \textbf{0.648} & \textbf{0.736} \\
            \midrule
            \multirow{7}{*}{$16\times$}
                & HiT-SR~\cite{zhang2024hierarchical} & 13.27 & 19.0 & 0.314 & 0.318 \\
                & MambaIR~\cite{guo2024mambair} & 13.42 & 19.8 & 0.315 & 0.334 \\
                & SwinIR~\cite{liang2021swinir} & 7.21 & 29.2 & 0.388 & 0.391 \\
                & SeeSR~\cite{wu2024seesr} & 6.49 & 53.0 & 0.528 & 0.535 \\
                & OSEDiff~\cite{wu2024one} & \textbf{5.79} & 53.5 & 0.550 & 0.611 \\
                & CoZ~\cite{kim2025chain} & 6.32 & \underline{58.9} & \underline{0.602} & \underline{0.663} \\
            \rowcolor{oursbg}
                & \textbf{\method{}} & \underline{5.85} & \textbf{61.6} & \textbf{0.605} & \textbf{0.735} \\
            \midrule
            \multirow{7}{*}{$64\times$}
                & HiT-SR~\cite{zhang2024hierarchical} & 16.55 & 22.0 & 0.370 & 0.408 \\
                & MambaIR~\cite{guo2024mambair} & 17.09 & 21.8 & 0.369 & 0.418 \\
                & SwinIR~\cite{liang2021swinir} & 8.62 & 22.9 & 0.474 & 0.460 \\
                & SeeSR~\cite{wu2024seesr} & 8.88 & 43.8 & 0.509 & 0.524 \\
                & OSEDiff~\cite{wu2024one} & \textbf{7.04} & 47.7 & 0.537 & 0.580 \\
                & CoZ~\cite{kim2025chain} & 7.78 & \underline{52.0} & \underline{0.584} & \underline{0.630} \\
            \rowcolor{oursbg}
                & \textbf{\method{}} & \underline{7.23} & \textbf{55.6} & \textbf{0.597} & \textbf{0.726} \\
            \midrule
            \multirow{7}{*}{$256\times$}
                & HiT-SR~\cite{zhang2024hierarchical} & 17.52 & 26.3 & 0.419 & 0.487 \\
                & MambaIR~\cite{guo2024mambair} & 18.43 & 26.4 & 0.416 & 0.497 \\
                & SwinIR~\cite{liang2021swinir} & 10.01 & 27.3 & 0.504 & 0.473 \\
                & SeeSR~\cite{wu2024seesr} & 10.79 & 37.6 & 0.505 & 0.510 \\
                & OSEDiff~\cite{wu2024one} & \underline{8.25} & 44.0 & 0.528 & 0.551 \\
                & CoZ~\cite{kim2025chain} & 8.83 & \underline{48.6} & \underline{0.577} & \underline{0.608} \\
            \rowcolor{oursbg}
                & \textbf{\method{}} & \textbf{8.07} & \textbf{51.8} & \textbf{0.593} & \textbf{0.710} \\
            \bottomrule
        \end{tabular}%
    }
    \caption{\textbf{Per-scale no-reference quality on DIV8K.}
    \textbf{Bold}: best. \underline{Underline}: second-best.}
    \label{tab:ds_div8k}
\end{table}

\begin{table}[t]
    \centering
    \small
    \renewcommand{\arraystretch}{0.9}
    \setlength{\tabcolsep}{4pt}
    \resizebox{\columnwidth}{!}{%
        \begin{tabular}{ll cccc}
            \toprule
            Scale & Method & NIQE$\downarrow$ & MUSIQ$\uparrow$ & MANIQA$\uparrow$ & CLIPIQA$\uparrow$ \\
            \midrule
            \multirow{7}{*}{$4\times$}
                & HiT-SR~\cite{zhang2024hierarchical} & 8.72 & 41.0 & 0.429 & 0.377 \\
                & MambaIR~\cite{guo2024mambair} & 8.63 & 42.8 & 0.446 & 0.382 \\
                & SwinIR~\cite{liang2021swinir} & 6.51 & 54.6 & 0.542 & 0.482 \\
                & SeeSR~\cite{wu2024seesr} & 5.85 & 66.9 & 0.640 & 0.639 \\
                & OSEDiff~\cite{wu2024one} & \textbf{5.59} & 59.3 & 0.587 & 0.636 \\
                & CoZ~\cite{kim2025chain} & 6.14 & \textbf{67.2} & \textbf{0.642} & \underline{0.670} \\
            \rowcolor{oursbg}
                & \textbf{\method{}} & \underline{5.71} & \underline{67.1} & \underline{0.642} & \textbf{0.704} \\
            \midrule
            \multirow{7}{*}{$16\times$}
                & HiT-SR~\cite{zhang2024hierarchical} & 13.37 & 19.5 & 0.321 & 0.354 \\
                & MambaIR~\cite{guo2024mambair} & 13.73 & 20.6 & 0.322 & 0.375 \\
                & SwinIR~\cite{liang2021swinir} & 7.37 & 35.9 & 0.398 & 0.435 \\
                & SeeSR~\cite{wu2024seesr} & 7.34 & 57.2 & 0.540 & 0.583 \\
                & OSEDiff~\cite{wu2024one} & \textbf{6.57} & 52.3 & 0.523 & 0.608 \\
                & CoZ~\cite{kim2025chain} & 7.28 & \underline{59.2} & \underline{0.574} & \underline{0.649} \\
            \rowcolor{oursbg}
                & \textbf{\method{}} & \underline{6.63} & \textbf{61.9} & \textbf{0.580} & \textbf{0.739} \\
            \midrule
            \multirow{7}{*}{$64\times$}
                & HiT-SR~\cite{zhang2024hierarchical} & 16.60 & 22.4 & 0.359 & 0.420 \\
                & MambaIR~\cite{guo2024mambair} & 17.20 & 22.0 & 0.359 & 0.432 \\
                & SwinIR~\cite{liang2021swinir} & 8.43 & 26.4 & 0.466 & 0.462 \\
                & SeeSR~\cite{wu2024seesr} & 9.31 & 48.0 & 0.514 & 0.553 \\
                & OSEDiff~\cite{wu2024one} & \underline{7.99} & 46.0 & 0.525 & 0.582 \\
                & CoZ~\cite{kim2025chain} & 8.22 & \underline{53.1} & \underline{0.563} & \underline{0.634} \\
            \rowcolor{oursbg}
                & \textbf{\method{}} & \textbf{7.29} & \textbf{55.7} & \textbf{0.580} & \textbf{0.735} \\
            \midrule
            \multirow{7}{*}{$256\times$}
                & HiT-SR~\cite{zhang2024hierarchical} & 17.95 & 25.1 & 0.411 & 0.477 \\
                & MambaIR~\cite{guo2024mambair} & 18.78 & 24.8 & 0.409 & 0.492 \\
                & SwinIR~\cite{liang2021swinir} & 10.51 & 27.1 & 0.487 & 0.459 \\
                & SeeSR~\cite{wu2024seesr} & 11.94 & 36.6 & 0.490 & 0.505 \\
                & OSEDiff~\cite{wu2024one} & \underline{9.03} & 41.8 & 0.522 & 0.542 \\
                & CoZ~\cite{kim2025chain} & 9.27 & \underline{48.1} & \underline{0.572} & \underline{0.590} \\
            \rowcolor{oursbg}
                & \textbf{\method{}} & \textbf{8.04} & \textbf{52.8} & \textbf{0.595} & \textbf{0.709} \\
            \bottomrule
        \end{tabular}%
    }
    \caption{\textbf{Per-scale no-reference quality on DRealSR.}
    \textbf{Bold}: best. \underline{Underline}: second-best.}
    \label{tab:ds_drealsr}
\end{table}

\begin{table}[t]
    \centering
    \small
    \renewcommand{\arraystretch}{0.9}
    \setlength{\tabcolsep}{4pt}
    \resizebox{\columnwidth}{!}{%
        \begin{tabular}{ll cccc}
            \toprule
            Scale & Method & NIQE$\downarrow$ & MUSIQ$\uparrow$ & MANIQA$\uparrow$ & CLIPIQA$\uparrow$ \\
            \midrule
            \multirow{7}{*}{$4\times$}
                & HiT-SR~\cite{zhang2024hierarchical} & 9.55 & 42.4 & 0.407 & 0.542 \\
                & MambaIR~\cite{guo2024mambair} & 9.44 & 42.8 & 0.407 & 0.561 \\
                & SwinIR~\cite{liang2021swinir} & 5.56 & 52.1 & 0.503 & 0.570 \\
                & SeeSR~\cite{wu2024seesr} & \underline{4.79} & \underline{65.6} & 0.597 & 0.709 \\
                & OSEDiff~\cite{wu2024one} & \textbf{4.76} & 64.2 & 0.594 & 0.726 \\
                & CoZ~\cite{kim2025chain} & 5.22 & 65.6 & \textbf{0.616} & \underline{0.741} \\
            \rowcolor{oursbg}
                & \textbf{\method{}} & 5.48 & \textbf{66.8} & \underline{0.611} & \textbf{0.777} \\
            \midrule
            \multirow{7}{*}{$16\times$}
                & HiT-SR~\cite{zhang2024hierarchical} & 15.95 & 22.2 & 0.327 & 0.358 \\
                & MambaIR~\cite{guo2024mambair} & 16.11 & 21.7 & 0.325 & 0.362 \\
                & SwinIR~\cite{liang2021swinir} & \underline{6.93} & 27.8 & 0.482 & 0.478 \\
                & SeeSR~\cite{wu2024seesr} & 7.76 & 43.8 & 0.509 & 0.523 \\
                & OSEDiff~\cite{wu2024one} & \textbf{6.40} & \underline{50.4} & 0.556 & 0.536 \\
                & CoZ~\cite{kim2025chain} & 7.46 & 47.3 & \underline{0.561} & \underline{0.541} \\
            \rowcolor{oursbg}
                & \textbf{\method{}} & 7.01 & \textbf{51.2} & \textbf{0.584} & \textbf{0.692} \\
            \midrule
            \multirow{7}{*}{$64\times$}
                & HiT-SR~\cite{zhang2024hierarchical} & 17.93 & 29.2 & 0.393 & 0.508 \\
                & MambaIR~\cite{guo2024mambair} & 18.60 & 28.3 & 0.390 & 0.518 \\
                & SwinIR~\cite{liang2021swinir} & 8.16 & 35.1 & 0.514 & 0.470 \\
                & SeeSR~\cite{wu2024seesr} & 10.59 & 39.5 & 0.506 & 0.514 \\
                & OSEDiff~\cite{wu2024one} & \textbf{6.92} & \underline{53.0} & 0.554 & 0.516 \\
                & CoZ~\cite{kim2025chain} & 7.78 & 50.2 & \underline{0.572} & \underline{0.554} \\
            \rowcolor{oursbg}
                & \textbf{\method{}} & \underline{7.74} & \textbf{54.7} & \textbf{0.606} & \textbf{0.736} \\
            \midrule
            \multirow{7}{*}{$256\times$}
                & HiT-SR~\cite{zhang2024hierarchical} & 17.85 & 29.9 & 0.438 & 0.523 \\
                & MambaIR~\cite{guo2024mambair} & 19.13 & 29.8 & 0.433 & 0.530 \\
                & SwinIR~\cite{liang2021swinir} & 9.20 & 34.5 & 0.513 & 0.467 \\
                & SeeSR~\cite{wu2024seesr} & 12.66 & 38.6 & 0.506 & 0.541 \\
                & OSEDiff~\cite{wu2024one} & \textbf{7.35} & \underline{52.4} & 0.552 & 0.514 \\
                & CoZ~\cite{kim2025chain} & \underline{8.10} & 51.4 & \underline{0.585} & \underline{0.579} \\
            \rowcolor{oursbg}
                & \textbf{\method{}} & 8.78 & \textbf{56.0} & \textbf{0.617} & \textbf{0.771} \\
            \bottomrule
        \end{tabular}%
    }
    \caption{\textbf{Per-scale no-reference quality on FFHQ.}
    \textbf{Bold}: best. \underline{Underline}: second-best.}
    \label{tab:ds_ffhq}
\end{table}

\begin{table}[t]
    \centering
    \small
    \renewcommand{\arraystretch}{0.9}
    \setlength{\tabcolsep}{4pt}
    \resizebox{\columnwidth}{!}{%
        \begin{tabular}{ll cccc}
            \toprule
            Scale & Method & NIQE$\downarrow$ & MUSIQ$\uparrow$ & MANIQA$\uparrow$ & CLIPIQA$\uparrow$ \\
            \midrule
            \multirow{7}{*}{$4\times$}
                & HiT-SR~\cite{zhang2024hierarchical} & 7.95 & 36.4 & 0.402 & 0.362 \\
                & MambaIR~\cite{guo2024mambair} & 7.88 & 37.6 & 0.418 & 0.367 \\
                & SwinIR~\cite{liang2021swinir} & 5.74 & 49.0 & 0.513 & 0.427 \\
                & SeeSR~\cite{wu2024seesr} & \underline{4.62} & 63.2 & 0.611 & 0.590 \\
                & OSEDiff~\cite{wu2024one} & 4.81 & 59.5 & 0.587 & 0.620 \\
                & CoZ~\cite{kim2025chain} & 4.64 & \textbf{66.8} & \underline{0.642} & \underline{0.685} \\
            \rowcolor{oursbg}
                & \textbf{\method{}} & \textbf{4.45} & \underline{66.4} & \textbf{0.645} & \textbf{0.722} \\
            \midrule
            \multirow{7}{*}{$16\times$}
                & HiT-SR~\cite{zhang2024hierarchical} & 13.12 & 17.8 & 0.306 & 0.301 \\
                & MambaIR~\cite{guo2024mambair} & 13.29 & 18.5 & 0.309 & 0.319 \\
                & SwinIR~\cite{liang2021swinir} & 6.78 & 27.1 & 0.378 & 0.375 \\
                & SeeSR~\cite{wu2024seesr} & 6.46 & 50.4 & 0.512 & 0.509 \\
                & OSEDiff~\cite{wu2024one} & \textbf{5.57} & 53.5 & 0.543 & 0.597 \\
                & CoZ~\cite{kim2025chain} & 6.08 & \underline{58.8} & \textbf{0.602} & \underline{0.656} \\
            \rowcolor{oursbg}
                & \textbf{\method{}} & \underline{5.63} & \textbf{61.9} & \underline{0.601} & \textbf{0.735} \\
            \midrule
            \multirow{7}{*}{$64\times$}
                & HiT-SR~\cite{zhang2024hierarchical} & 16.44 & 21.4 & 0.362 & 0.394 \\
                & MambaIR~\cite{guo2024mambair} & 17.01 & 21.2 & 0.362 & 0.403 \\
                & SwinIR~\cite{liang2021swinir} & 8.06 & 22.5 & 0.476 & 0.459 \\
                & SeeSR~\cite{wu2024seesr} & 8.40 & 42.5 & 0.500 & 0.511 \\
                & OSEDiff~\cite{wu2024one} & \textbf{6.72} & 48.4 & 0.542 & 0.571 \\
                & CoZ~\cite{kim2025chain} & 7.51 & \underline{52.8} & \underline{0.583} & \underline{0.629} \\
            \rowcolor{oursbg}
                & \textbf{\method{}} & \underline{6.85} & \textbf{57.0} & \textbf{0.598} & \textbf{0.730} \\
            \midrule
            \multirow{7}{*}{$256\times$}
                & HiT-SR~\cite{zhang2024hierarchical} & 17.50 & 26.1 & 0.417 & 0.479 \\
                & MambaIR~\cite{guo2024mambair} & 18.31 & 26.1 & 0.415 & 0.489 \\
                & SwinIR~\cite{liang2021swinir} & 9.05 & 28.4 & 0.510 & 0.454 \\
                & SeeSR~\cite{wu2024seesr} & 10.45 & 37.7 & 0.503 & 0.500 \\
                & OSEDiff~\cite{wu2024one} & \underline{8.02} & 45.7 & 0.540 & 0.556 \\
                & CoZ~\cite{kim2025chain} & 8.40 & \underline{50.5} & \underline{0.582} & \underline{0.602} \\
            \rowcolor{oursbg}
                & \textbf{\method{}} & \textbf{7.73} & \textbf{54.1} & \textbf{0.599} & \textbf{0.719} \\
            \bottomrule
        \end{tabular}%
    }
    \caption{\textbf{Per-scale no-reference quality on Flickr2K.}
    \textbf{Bold}: best. \underline{Underline}: second-best.}
    \label{tab:ds_flickr2k}
\end{table}

\begin{table}[t]
    \centering
    \small
    \renewcommand{\arraystretch}{0.9}
    \setlength{\tabcolsep}{4pt}
    \resizebox{\columnwidth}{!}{%
        \begin{tabular}{ll cccc}
            \toprule
            Scale & Method & NIQE$\downarrow$ & MUSIQ$\uparrow$ & MANIQA$\uparrow$ & CLIPIQA$\uparrow$ \\
            \midrule
            \multirow{7}{*}{$4\times$}
                & HiT-SR~\cite{zhang2024hierarchical} & 9.22 & 34.7 & 0.372 & 0.346 \\
                & MambaIR~\cite{guo2024mambair} & 9.29 & 35.0 & 0.381 & 0.364 \\
                & SwinIR~\cite{liang2021swinir} & 6.78 & 50.4 & 0.510 & 0.447 \\
                & SeeSR~\cite{wu2024seesr} & 6.41 & 61.1 & 0.588 & 0.557 \\
                & OSEDiff~\cite{wu2024one} & \textbf{5.83} & 57.7 & 0.578 & 0.612 \\
                & CoZ~\cite{kim2025chain} & 6.18 & \textbf{64.3} & \textbf{0.619} & \underline{0.643} \\
            \rowcolor{oursbg}
                & \textbf{\method{}} & \underline{5.95} & \underline{63.5} & \underline{0.602} & \textbf{0.687} \\
            \midrule
            \multirow{7}{*}{$16\times$}
                & HiT-SR~\cite{zhang2024hierarchical} & 13.95 & 19.5 & 0.320 & 0.360 \\
                & MambaIR~\cite{guo2024mambair} & 14.45 & 19.6 & 0.322 & 0.377 \\
                & SwinIR~\cite{liang2021swinir} & 7.20 & 33.3 & 0.422 & 0.439 \\
                & SeeSR~\cite{wu2024seesr} & 7.70 & 52.5 & 0.527 & 0.533 \\
                & OSEDiff~\cite{wu2024one} & \textbf{6.75} & 53.7 & 0.543 & 0.603 \\
                & CoZ~\cite{kim2025chain} & 7.60 & \underline{57.3} & \underline{0.578} & \underline{0.626} \\
            \rowcolor{oursbg}
                & \textbf{\method{}} & \underline{6.92} & \textbf{59.4} & \textbf{0.580} & \textbf{0.702} \\
            \midrule
            \multirow{7}{*}{$64\times$}
                & HiT-SR~\cite{zhang2024hierarchical} & 16.65 & 23.8 & 0.387 & 0.457 \\
                & MambaIR~\cite{guo2024mambair} & 17.30 & 23.5 & 0.387 & 0.470 \\
                & SwinIR~\cite{liang2021swinir} & 8.31 & 25.6 & 0.484 & 0.470 \\
                & SeeSR~\cite{wu2024seesr} & 10.34 & 44.2 & 0.499 & 0.513 \\
                & OSEDiff~\cite{wu2024one} & \textbf{7.64} & 47.0 & 0.527 & 0.557 \\
                & CoZ~\cite{kim2025chain} & 8.66 & \underline{50.0} & \underline{0.560} & \underline{0.585} \\
            \rowcolor{oursbg}
                & \textbf{\method{}} & \underline{7.86} & \textbf{54.0} & \textbf{0.575} & \textbf{0.708} \\
            \midrule
            \multirow{7}{*}{$256\times$}
                & HiT-SR~\cite{zhang2024hierarchical} & 17.58 & 26.2 & 0.429 & 0.479 \\
                & MambaIR~\cite{guo2024mambair} & 18.67 & 25.9 & 0.429 & 0.495 \\
                & SwinIR~\cite{liang2021swinir} & 10.78 & 28.4 & 0.497 & 0.454 \\
                & SeeSR~\cite{wu2024seesr} & 12.27 & 34.2 & 0.495 & 0.490 \\
                & OSEDiff~\cite{wu2024one} & \textbf{8.39} & 43.1 & 0.521 & 0.517 \\
                & CoZ~\cite{kim2025chain} & 9.83 & \underline{45.2} & \underline{0.556} & \underline{0.540} \\
            \rowcolor{oursbg}
                & \textbf{\method{}} & \underline{8.72} & \textbf{49.1} & \textbf{0.585} & \textbf{0.667} \\
            \bottomrule
        \end{tabular}%
    }
    \caption{\textbf{Per-scale no-reference quality on RealSR.}
    \textbf{Bold}: best. \underline{Underline}: second-best.}
    \label{tab:ds_realsr}
\end{table}

\vspace{3pt}
\noindent\textbf{Quantitative Results.}
\Cref{tab:ds_4klsdb,tab:ds_div2k,tab:ds_div8k,tab:ds_drealsr,tab:ds_ffhq,tab:ds_flickr2k,tab:ds_realsr} report the complete per-scale no-reference results on all seven test sets.
\method{} achieves the highest CLIPIQA at every scale on every dataset.
At $4{\times}$, MUSIQ and MANIQA often favor another method.
This $4{\times}$ pattern is consistent with the target-available objective, which emphasizes agreement with the available ground truth.
Beyond the supervision boundary, the gap between \method{} and the other methods becomes clearer.
From $16{\times}$ onward, \method{} achieves the best MUSIQ and MANIQA on nearly every dataset-scale pair.
The only exception is MANIQA on Flickr2K at $16{\times}$.
This trend holds on the high-resolution photographic datasets and also on the real-camera benchmarks DRealSR and RealSR.
The differences are most pronounced at $64{\times}$ and $256{\times}$, after several recursive SR steps.
At $256{\times}$, \method{} exceeds CoZ in CLIPIQA by more than $0.10$ on all seven datasets.
The margin ranges from $0.102$ on DIV8K to $0.192$ on FFHQ.
This margin is larger than the margin at $16{\times}$, the first scale beyond the supervision boundary, and it grows with zoom depth.
The trend is also consistent across very different image content, including natural photographs, faces, real-camera images, and high-resolution texture-rich scenes.
NIQE gives mixed rankings.
In contrast, the learned no-reference metrics (MUSIQ, MANIQA, and CLIPIQA) consistently favor \method{} at deeper scales.
The result holds across several metrics and datasets.
The strongest gains appear repeatedly where recursive errors accumulate the most.
The per-dataset results support the main finding that the advantage of \method{} becomes clearer as recursion proceeds beyond direct supervision.

\begin{figure*}[t]
\centering
\includegraphics[width=0.9\linewidth]{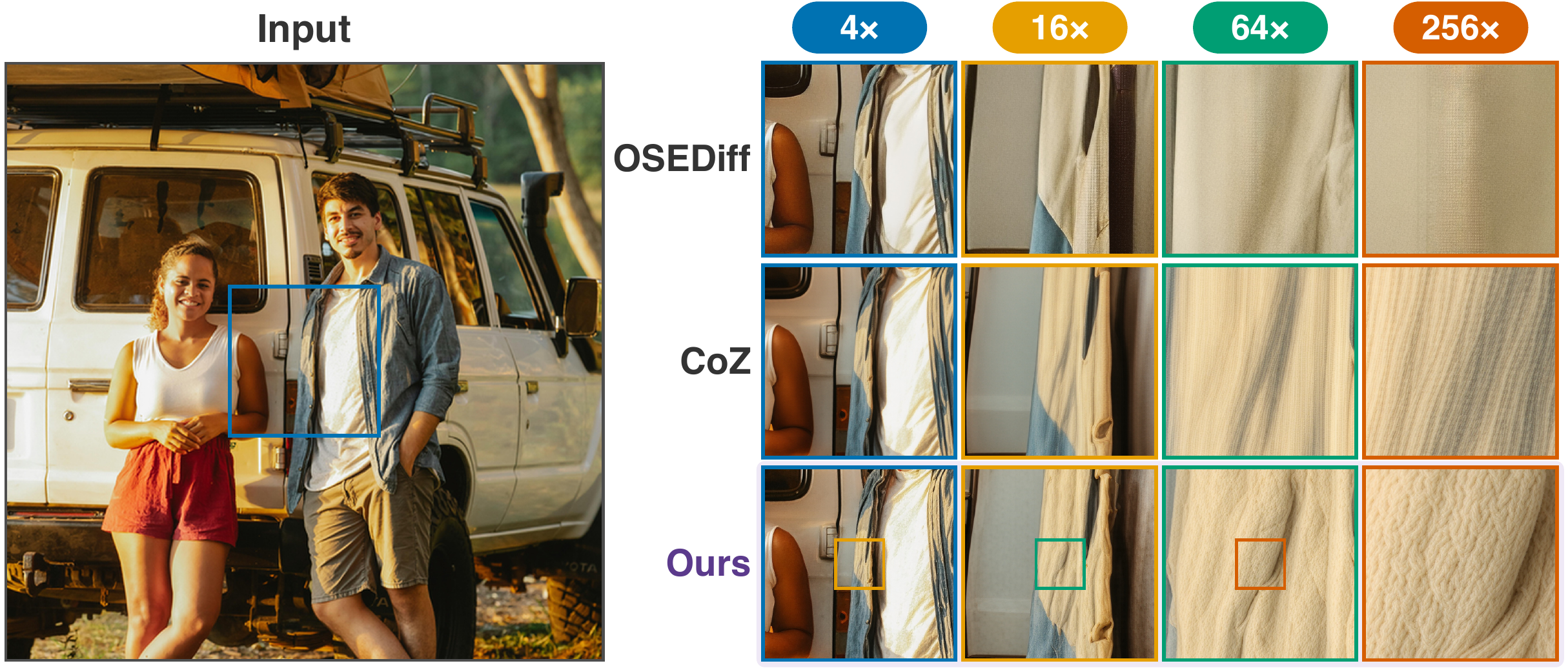}
\caption{
\textbf{Qualitative results on 4KLSDB from $4\times$ to $256\times$.}
The zoom follows a waffle-knit garment.
OSEDiff flattens the fabric into a uniform field with a grid-like texture, and CoZ stretches the folds into vertical bands.
\method{} keeps the raised folds and the interlocking knit.
Colored boxes mark the region enlarged at the next step.
}
\label{fig:ds_4klsdb}
\end{figure*}

\begin{figure*}[t]
\centering
\includegraphics[width=0.9\linewidth]{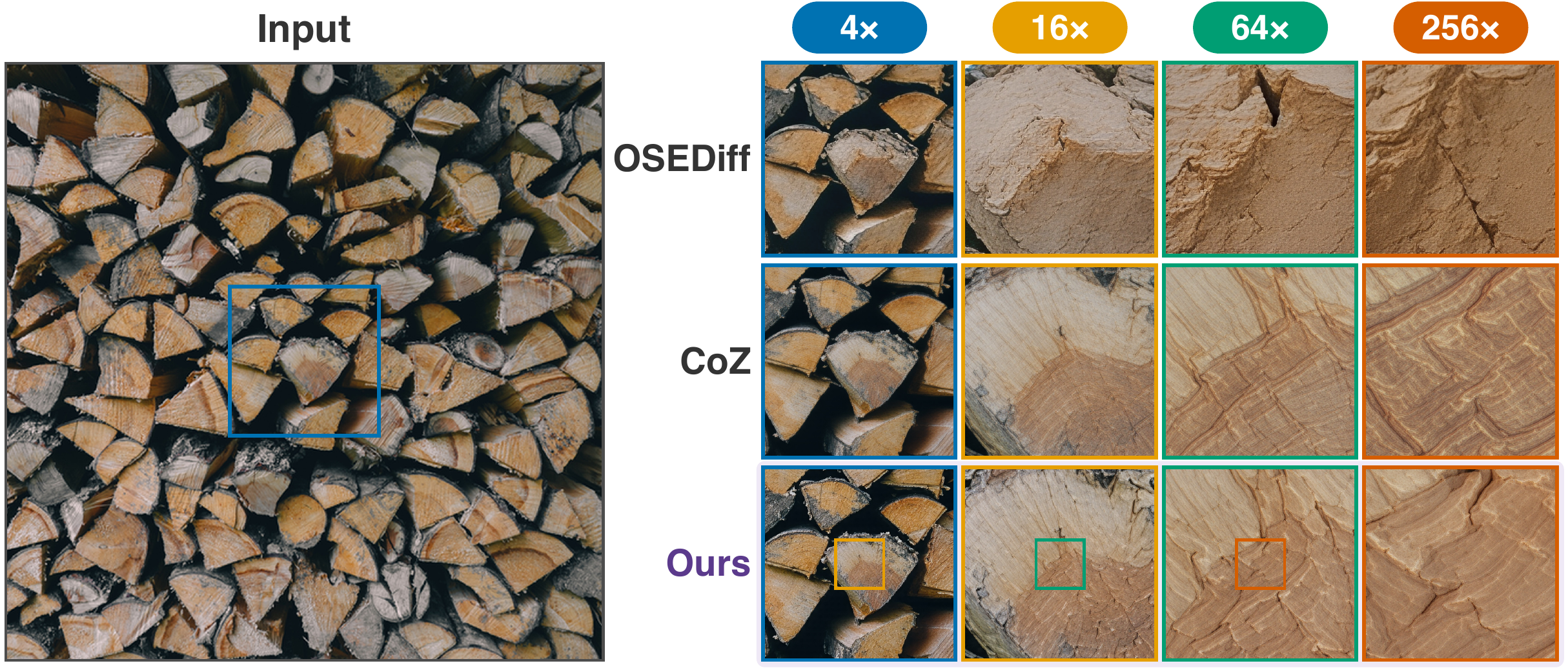}
\caption{
\textbf{Qualitative results on DIV2K from $4\times$ to $256\times$.}
The zoom follows the cut end of a log in a woodpile.
OSEDiff turns the wood into cracked, clay-like mud, and CoZ adds swirling, marble-like bands.
\method{} keeps growth rings and splits that look like wood up to $256\times$.
Colored boxes mark the region enlarged at the next step.
}
\label{fig:ds_div2k}
\end{figure*}

\begin{figure*}[t]
\centering
\includegraphics[width=0.9\linewidth]{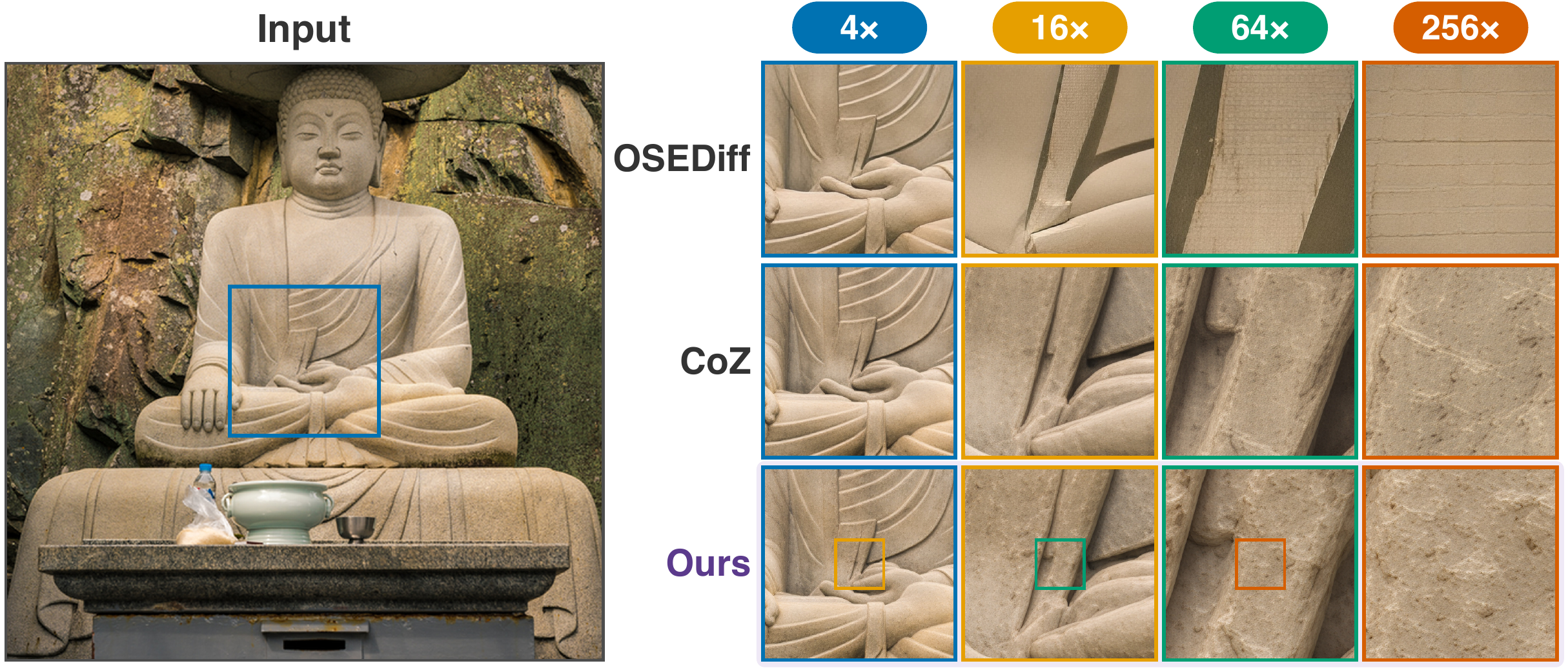}
\caption{
\textbf{Qualitative results on DIV8K from $4\times$ to $256\times$.}
The zoom follows the carved folds of a sandstone statue.
OSEDiff adds regular mesh-like patterns, and CoZ smooths the surface into broad waves.
\method{} keeps the carved relief and the irregular stone texture.
Colored boxes mark the region enlarged at the next step.
}
\label{fig:ds_div8k}
\end{figure*}

\begin{figure*}[t]
\centering
\includegraphics[width=0.9\linewidth]{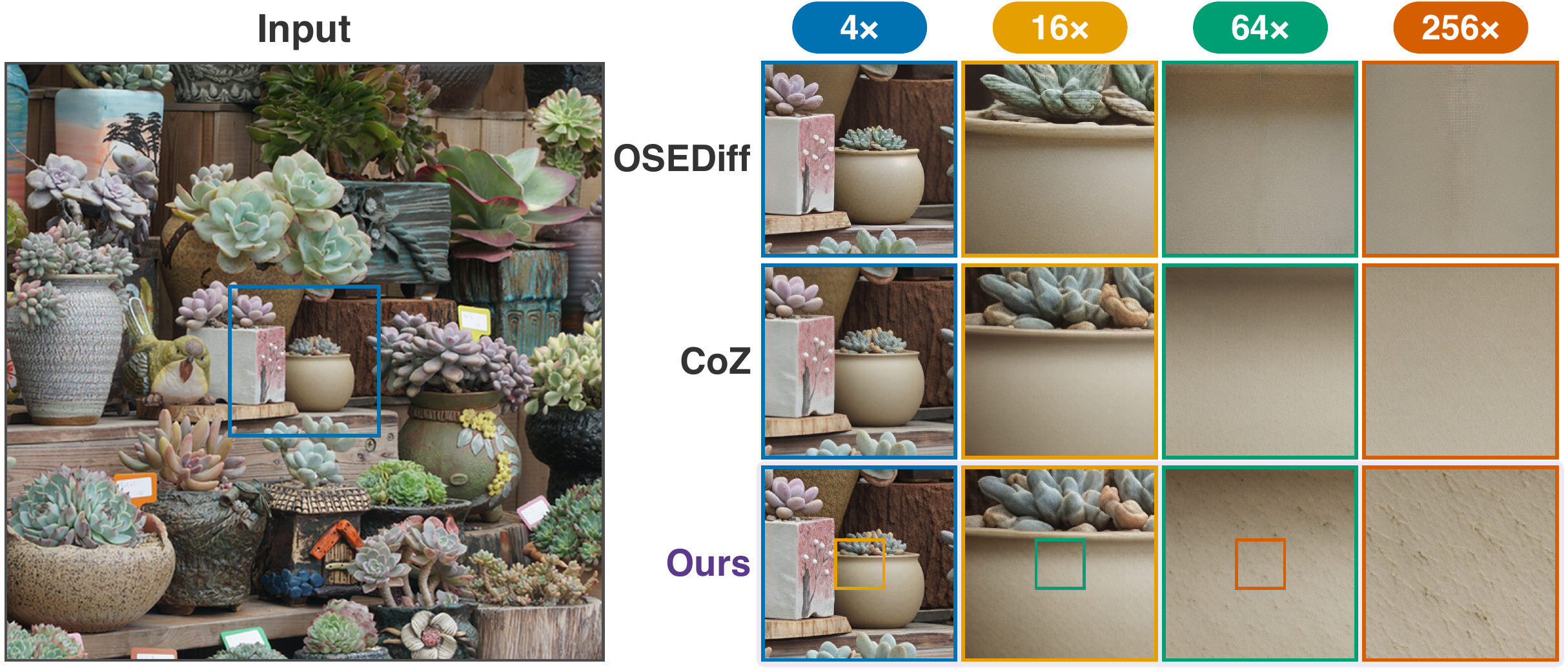}
\caption{
\textbf{Qualitative results on DRealSR from $4\times$ to $256\times$.}
The zoom follows the surface of an unglazed stoneware planter.
Both baselines lose the surface variation and become nearly flat at deeper scales.
\method{} keeps a granular texture, with varying direction, up to $256\times$.
Colored boxes mark the region enlarged at the next step.
}
\label{fig:ds_drealsr}
\end{figure*}

\begin{figure*}[t]
\centering
\includegraphics[width=0.9\linewidth]{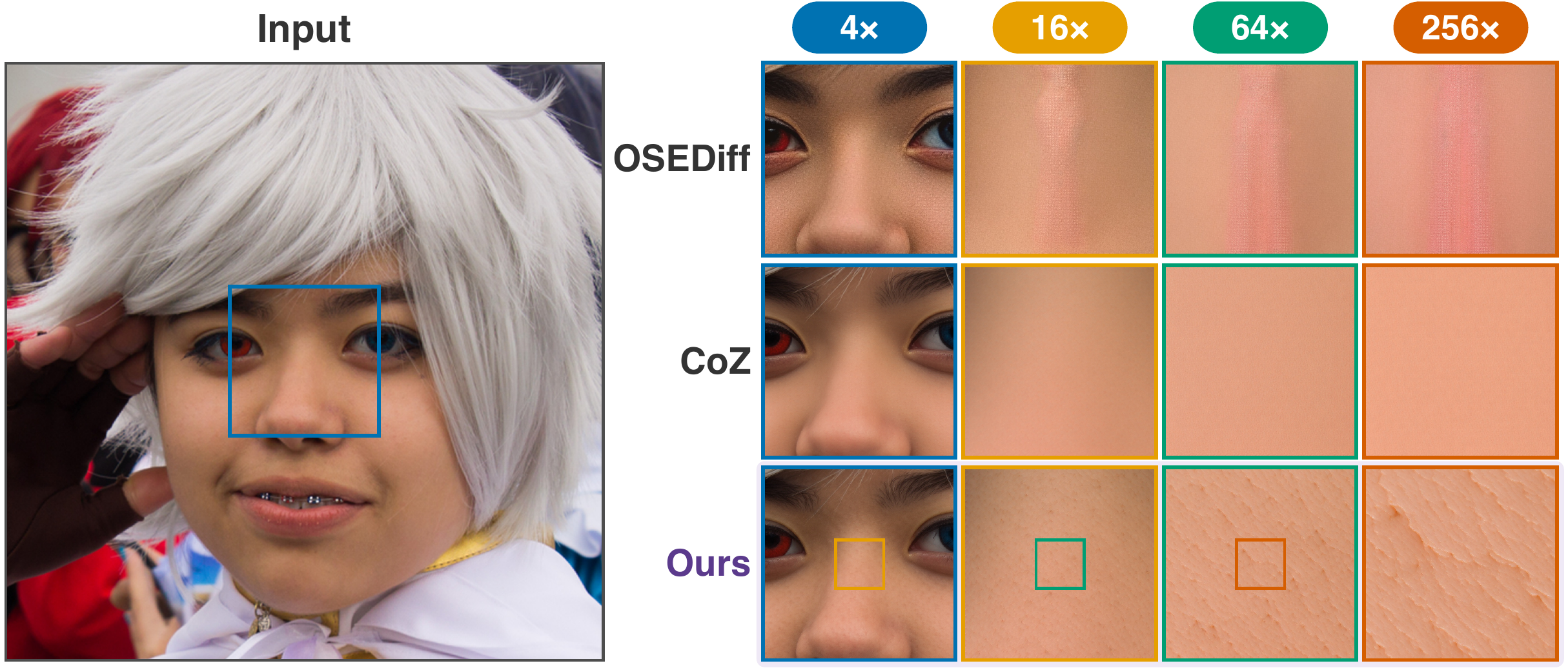}
\caption{
\textbf{Qualitative results on FFHQ from $4\times$ to $256\times$.}
The zoom moves from below the eyes onto the cheek.
OSEDiff adds a pink vertical streak that grows with depth, and CoZ smooths the skin into a flat field.
\method{} keeps fine skin creases up to $256\times$.
Colored boxes mark the region enlarged at the next step.
}
\label{fig:ds_ffhq}
\end{figure*}

\begin{figure*}[t]
\centering
\includegraphics[width=0.9\linewidth]{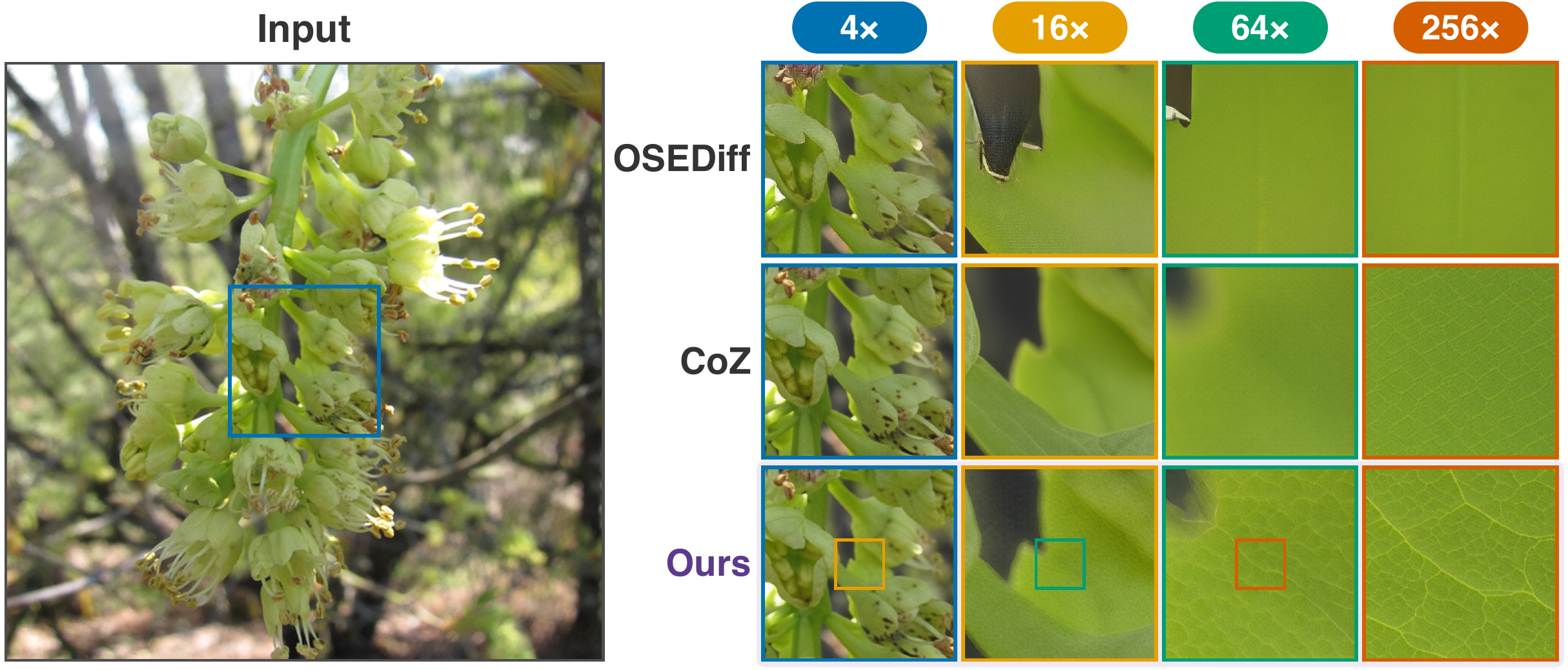}
\caption{
\textbf{Qualitative results on Flickr2K from $4\times$ to $256\times$.}
The zoom moves from a flower cluster onto a leaf.
OSEDiff flattens the leaf into a uniform green field with faint seams, and CoZ stays blurry until a uniform mesh appears at $256\times$.
\method{} recovers a connected network of leaf veins.
Colored boxes mark the region enlarged at the next step.
}
\label{fig:ds_flickr2k}
\end{figure*}

\begin{figure*}[t]
\centering
\includegraphics[width=0.9\linewidth]{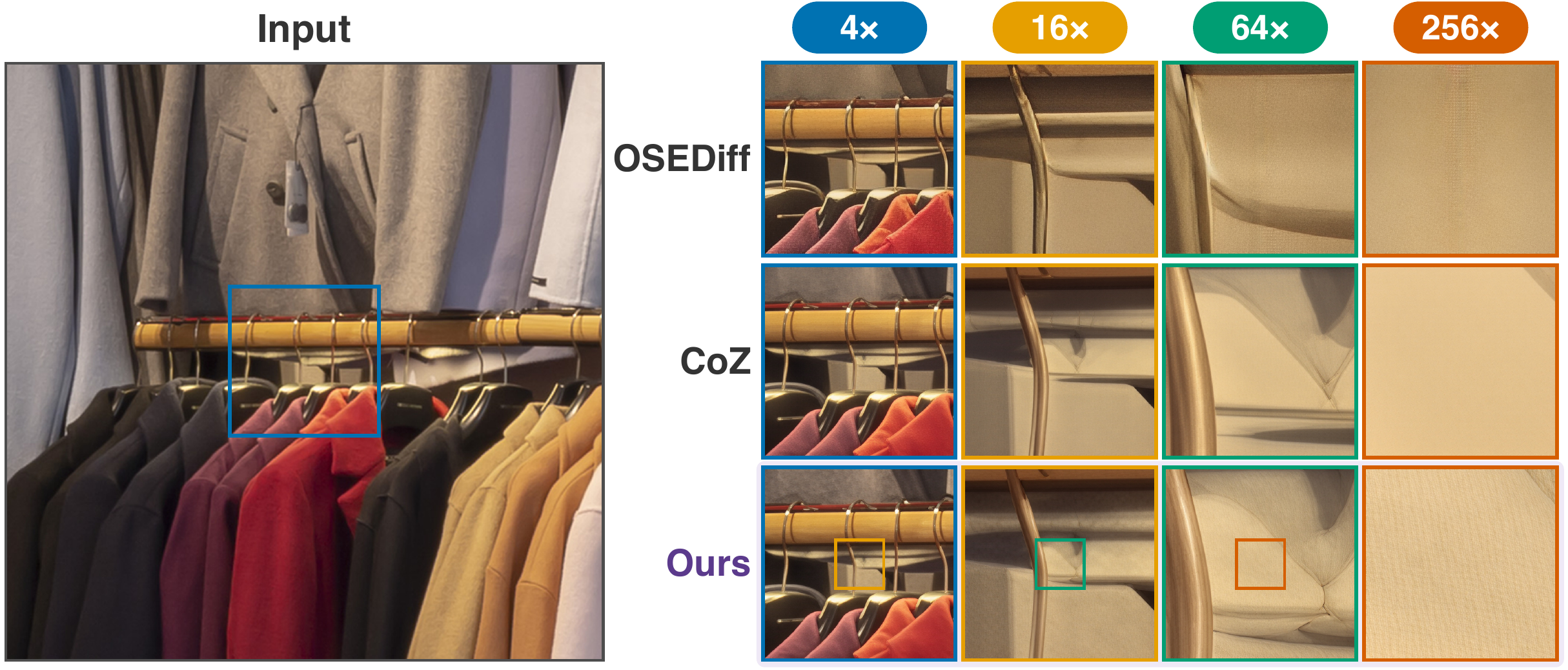}
\caption{
\textbf{Qualitative results on RealSR from $4\times$ to $256\times$.}
The zoom follows a garment hanging on a clothing rack.
OSEDiff and CoZ reduce the garment to smooth or regular patterns at deeper scales.
\method{} keeps more of the garment structure and fabric variation, most clearly at $256\times$.
Colored boxes mark the region enlarged at the next step.
}
\label{fig:ds_realsr}
\end{figure*}

\vspace{3pt}
\noindent\textbf{Qualitative Results.}
\Cref{fig:ds_4klsdb,fig:ds_div2k,fig:ds_div8k,fig:ds_drealsr,fig:ds_ffhq,fig:ds_flickr2k,fig:ds_realsr} show complete $4{\times}\!\rightarrow16{\times}\!\rightarrow64{\times}\!\rightarrow256{\times}$ recursive zooms across all seven datasets.
A consistent failure pattern emerges as recursion deepens.

\textbf{OSEDiff} often replaces the original material with regular synthetic textures or flat, uniform regions.
Examples are fine regular grids on knit fabric, cracked clay on cut wood, mesh patterns on sandstone, and color streaks on skin. \textbf{CoZ} preserves the coarse region more often than OSEDiff, but it progressively loses the material structure of that region.
For example, it turns wood grain into marble-like bands and stone into smooth surfaces that look like wax.
It also turns planters into nearly uniform regions, skin into a flat region, and a leaf into blur. In contrast, \textbf{\method{}} keeps more of the material structure of each region.
Examples are interlocking knit and folds in 4KLSDB, growth rings in DIV2K, and carved relief and stone grain in DIV8K.
Further examples are granular planter texture in DRealSR, leaf veins in Flickr2K, and fabric structure in RealSR.

\begin{figure*}[t]
\centering
\includegraphics[width=0.9\linewidth]{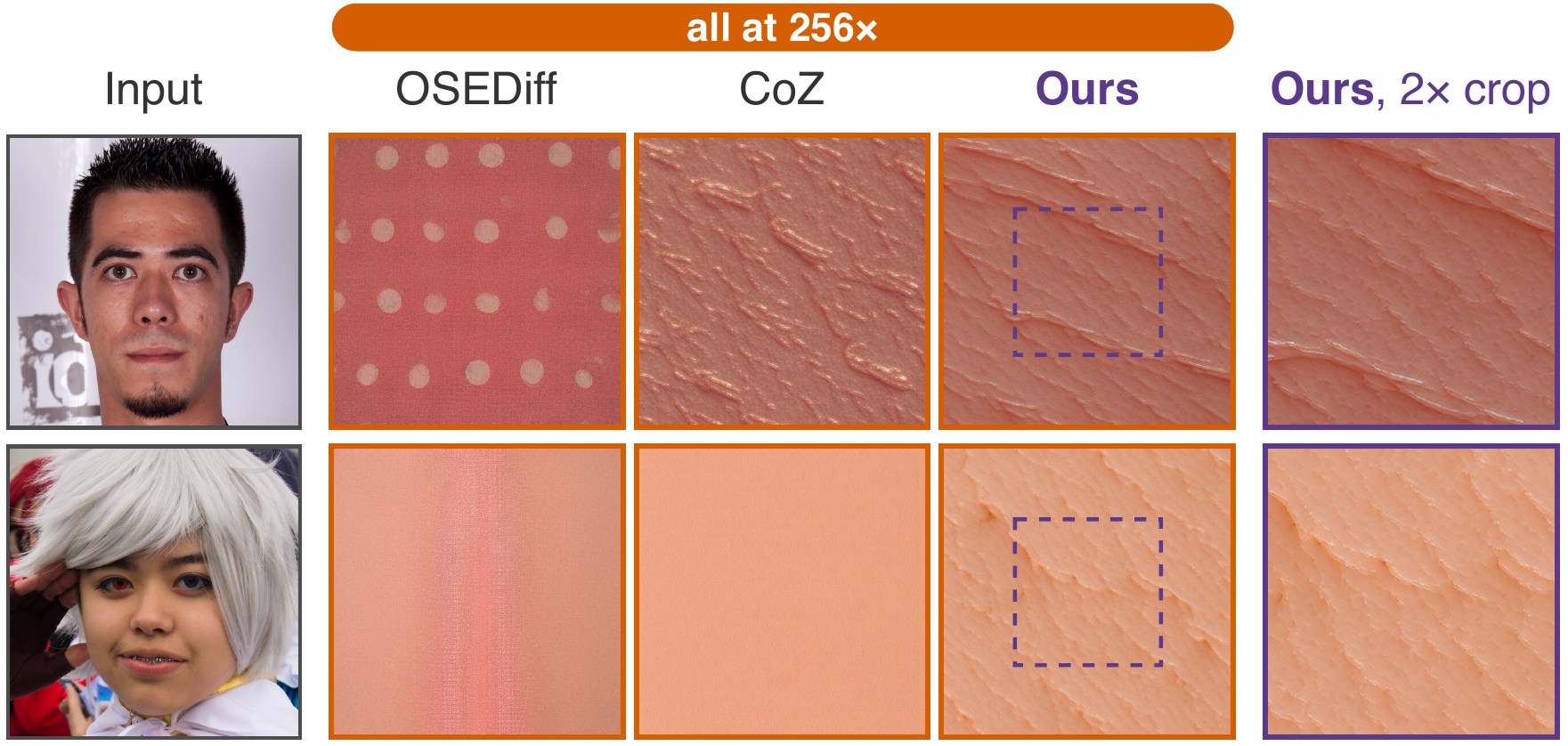}
\caption{
\textbf{Failure case: smooth skin at $256\times$.}
Here the projection to the last ground truth determines only the coarse shading, so the quality model shapes the pore-level texture (\cref{app:cross_scale_ambiguity}).
\method{} then draws crease-like ridges that are sharper and more regular than real skin, and the skin looks slightly artificial up close (right: the dashed region, enlarged $2\times$).
The baselines fail in other ways here: OSEDiff adds a dot grid or a color streak, and CoZ adds flake-like spots or flattens the skin.
}
\label{fig:failure_skin}
\end{figure*}

On structured materials such as knit fabric and carved stone, \method{} keeps repeated local geometry without turning it into an artificial grid.
On irregular natural surfaces such as wood and stoneware, \method{} keeps the nonuniform local variation.
The baselines make the region nearly smooth or regularly patterned. The difference is most visible at $64{\times}$ and $256{\times}$, where repeated use of earlier predictions amplifies small errors in the baselines.
The RealSR example is more challenging, and \method{} shows a smaller advantage at earlier scales.
However, the difference becomes clear at $256{\times}$.
At this scale, \method{} retains visible fabric structure, while both baselines become largely uniform.
On FFHQ, the learned no-reference metrics favor \method{}, but skin is its main failure case.\vadjust{\vfill\break}
At deeper zooms, it draws ridges that look sharper and more regular than real skin (\cref{fig:failure_skin}).
These examples show how the generated detail evolves as the zoom goes deeper.
Across the seven datasets, the qualitative evidence matches the trend of the no-reference metrics.
\Cref{app:limitations} reports the judge results on faces.
The methods are relatively close near the supervision boundary ($4{\times}$).
At deeper zooms, the differences between methods become increasingly visible.

\FloatBarrier

\FloatBarrier
\twocolumn[{\begin{minipage}{\textwidth}
\centering
\begin{tcolorbox}[enhanced, width=\linewidth, colback=tzBlueFill, colframe=tzBlueBorder, boxrule=1.2pt, arc=6pt, left=6pt, right=6pt, top=4pt, bottom=4pt,
    title={\small Ground-truth-anchored judges (Gemini 4 Argon and GPT-6.1 Sol), all scales}, coltitle=white, colbacktitle=tzBlueHeader2, fonttitle=\bfseries]
  \small\setlength{\parindent}{0pt}\raggedright
  You are judging the faithfulness of extreme image super-resolution. You get THREE images. Each has a LABEL printed in a black bar at its top-left: \texttt{'ANCHOR'}, \texttt{'A'}, or \texttt{'B'}. Use ONLY the printed label to identify an image.

  \smallskip
  ANCHOR is the REAL photograph of one small scene region. It has low resolution, so it can look blurry or blocky. A and B are two high-resolution reconstructions of exactly the same region, made by two methods.

  \smallskip
  A faithful reconstruction may add fine detail that the anchor is too blurry to show. It HALLUCINATES when it shows colors, large structures, edges, objects, or patterns that CONTRADICT what the anchor shows. Blur in the anchor is not evidence that the region is smooth. Judge consistency with the real anchor only, not which image looks nicer. Answer with JSON only:
  \begin{jsonbox}
    \ttfamily\footnotesize
    \{"saw": "anchor=..., A=..., B=...", "winner": "A" | "B" | "tie", "A\_hallucinates": true|false, "B\_hallucinates": true|false, "reason": "<one sentence>"\}
  \end{jsonbox}
\end{tcolorbox}
\captionof{figure}{\textbf{Prompt of the ground-truth-anchored judges.} The anchor is the native ground-truth window, the ground-truth region that the zoom step magnifies. Each pair is judged in both A/B orders.}
\label{fig:rubric_gemini}
\end{minipage}\vspace{14pt}}]

\section{Limitations}
\label{app:limitations}

Beyond $4\times$, projected DISTS (P-DISTS) and the two ground-truth-anchored judges (Gemini 4 Argon and GPT-6.1 Sol) measure consistency with the observable evidence.
Judges that compare each output with its previous zoom, instead of the ground truth, tend to favor smoother outputs.
The reason is that a lower-resolution image does not imply the added detail.
The current evaluation also uses synthetic recursion that always zooms into the center crop.
Physical camera zoom and user-selected regions remain for future work.
Training uses a curated set of 4K photographs.
We use only the high-resolution images of DRealSR and RealSR, so we do not test real-world degraded inputs, and fidelity on these two sets is mixed.
Faces remain the main failure case.
On FFHQ, the ground-truth-anchored judges flag \method{} more often than CoZ at $64\times$ and $256\times$ ($0.75$ vs.\ $0.18$ and $0.78$ vs.\ $0.29$).
At $256\times$, skin and other smooth organic surfaces can still look slightly artificial up close (\cref{fig:failure_skin}).
Finally, \method{} relies on a fixed no-reference quality model and on the frozen base model $F_0$ through the latent prior.
Both can bias the type of detail that training encourages at scales beyond the supervision boundary.
Future work should extend supervision to degradation models and adaptive zoom paths.
It should also develop stronger reference-aware evaluation for deeper recursive magnification.
\section{Judge Prompt}
\label{app:prompts}

\Cref{fig:rubric_gemini} gives the prompt of the ground-truth-anchored judges (Gemini 4 Argon and GPT-6.1 Sol).
Both judges receive the anchor and the two candidates as labeled images and answer in strict JSON.

\end{document}